\documentclass[11pt]{article}
\usepackage{caption}
\usepackage{subcaption}
\usepackage{amsfonts}
\usepackage{bbm}
\usepackage{amsmath}
\usepackage[ruled,vlined,linesnumbered]{algorithm2e}
\usepackage[section]{placeins}
\usepackage{epsfig}
\usepackage{amssymb}
\usepackage[numbers]{natbib}
\usepackage{graphicx}
\usepackage{amsthm}
\usepackage{booktabs}
\usepackage{etoolbox}
\usepackage[round-mode=places,detect-weight=true,detect-inline-weight=math]{siunitx}

\theoremstyle{definition}

\begin{document}

\title{A Vision Inspired Neural Network for Unsupervised Anomaly Detection in Unordered Data}

\author{Nassir Mohammad \\ \emph{Cyber Innovation and Scouting, Airbus, Newport, UK}}

\maketitle



\begin{abstract}
A fundamental problem in the field of unsupervised machine learning is the detection of anomalies corresponding to rare and unusual observations of interest; reasons include for their rejection, accommodation or further investigation. Anomalies are intuitively understood to be something unusual or inconsistent, whose occurrence sparks immediate attention. More formally anomalies are those observations---under appropriate random variable modelling---whose expectation of occurrence with respect to a grouping of prior interest is less than one; such a definition and understanding has been used to develop the parameter-free perception anomaly detection algorithm. The present work seeks to establish important and practical connections between the approach used by the perception algorithm and prior decades of research in neurophysiology and computational neuroscience; particularly that of information processing in the retina and visual cortex. The algorithm is conceptualised as a neuron model which forms the kernel of an unsupervised neural network that learns to signal unexpected observations as anomalies. Both the network and neuron display properties observed in biological processes including: immediate intelligence; parallel processing; redundancy; global degradation; contrast invariance; parameter-free computation, dynamic thresholds and non-linear processing. A robust and accurate model for anomaly detection in univariate and multivariate data is built using this network as a concrete application.
\end{abstract}


%
\section{Introduction} 
\label{section:introduction} 
%

The material and immaterial pursuit of understanding the representation and processes of the human mind and brain has resulted in a number of theories. One such system of ideas of interest to the present work is that of connectionism, otherwise known as artificial neural networks (ANNs). In this work neurophysiological simplifications of the representation and processes carried out by the human brain have been used to build models of networked computational units for solving `intelligent' tasks by learning through experience and without explicit programming. The assumptions and idealisations are many; taking some of these, along with observations carried out in the mammalian (and amphibian) retina and visual cortex, they are used in the present work as inspiration and the basis of a neural network for the perception of anomalies corresponding to meaningful information. 

Anomaly (or outlier) detection finds application in many domains including cyber security, medicine, machine vision, statistics, neuroscience, law enforcement and financial fraud---to name only a few. Anomalies were initially searched for clear rejection or omission from the data to aid statistical analysis, for example to compute the mean or standard deviation. They were also removed to better predictions from models such as linear regression, and more recently their removal aids the performance of machine learning algorithms. However, in many applications anomalies themselves are of interest and are the observations most desirous in the entire data set---which need to be identified and separated from noise or irrelevant outliers. 

The objective of anomaly detection algorithms is to detect rare, unusual or inconsistent observations from the rest of the data. It has been a historically important problem for which solutions have developed from early intuitive analysis, to statistical techniques and more recently to using computational approaches. Anomaly detection was originally carried out on small datasets using human intuition developed through domain knowledge and experience but without formal methods. This gradually gave way to statistical methods that were rapidly developing in the early and mid twentieth century almost always under the assumption that data were Gaussian distributed. This assumption was for theoretical convenience, ease of analysis, results about the Central Limit Theorem and as a carry on from past dogmas \cite{Hampel2001}. However, many real world distributions are not Gaussian and hence many of the popular outlier detection tests do not have their assumptions met by the data, yet continue to be used throughout the sciences. To address the requirements of processing non-Gaussian and large datasets the increase in computational power in recent decades has enabled the development of heuristic based and compute intensive approaches where practical results are prioritised over mathematically derived and well understood solutions. These being mainly non-parametric are suitable to a much wider range of data but do not necessarily outperform statistical approaches which have also benefited from the increased computational power. 

Most anomaly detection solutions optimise modelling the normal data so that observations that increasingly do not fit the model of normality are considered increasingly anomalous, while a small handful of solutions directly isolate anomalies. However, as has been the case for traditional statistical approaches, users are almost always left with the problem of deriving and justifying critical parameters which are often data specific in unsupervised learning problems.

Anomaly detection has always been considered to be a difficult problem due to its subjectivity. \citet{Barnett1978} clearly stated in their classic book on the subject that the major problem in outlier study remains even after surveying the vast literature: ``It is a matter of subjective judgement on the part of the observer whether or not he picks out some observation (or set of observations) for scrutiny \dots what characterises the `outlier' is its impact on the observer (it appears extreme in some way) \dots when all is said and done, the major problem in outlier study remains the one that faced the very earliest workers in the subject---\emph{what is an outlier?} We have taken the view that the stimulus lies in the subjective concept of surprise engendered by one, or a few, observations in a set of data \dots (and that) the concept is a human one". \citet{Barnett1978} foresaw the difficulty of translating the problem into a mechanised form and that ``trying to teach the computer what is surprising is difficult". 

Many definitions of an anomaly have been given with vaguely defined terms such as `surprising', `discordant', `inconsistent' and `discrepant' attempting to link the subjective human experience to objective tests for their detection. Unsatisfied with the nebulous descriptions, parameter laden approaches and the difficulties encountered in practical data analysis, \citet{nassir2021anomaly} took a human centric non-parametric approach to anomaly detection inspired by human visual perception and the Gestalt Theory of Psychology to provide the following definition of an anomaly that is abided by in the present work: 
\\

\noindent \emph{A grouping of interest represented by a gestalt law is perceived by the Helmholtz principle when it is unexpected to happen (i.e., its expectation of occurrence is $<1$) in uniform random noise. Any observation that is unexpected to occur with respect to this grouping is perceived, by the same principle, to be an anomaly.}
\\

Under this definition and understanding \citet{nassir2021anomaly} developed the unsupervised anomaly detection algorithm called the \emph{perception} for univariate and multivariate data using principles of human perception; namely Wertheimer's Contrast Invariance, Isotropy and the Helmholtz principle \citep{Morel07}. In addition, the algorithm utilised elements from the study of anomaly detection by various communities in computer vision, statistics, machine learning and data mining; thus developing a vision inspired, statistically founded, machine learning algorithm capable of handling the computational requirements of modern day data mining problems. The overall approach was motivated by the desire to have a method that is not only derived from sound principles but also tries to follow the mode of human perception in detecting anomalies which is both fast and accurate even on complex data, and to provide a parameter-free anomaly detection algorithm that is simple to implement, accurate, efficient and adaptable to the data distribution. In practice the algorithm has extremely fast training and prediction times, and performs relatively well in both subjective measurements and on real-world \emph{artificially preprocessed} anomaly detection data sets. Although the current version assumes anomalies are \emph{global} in nature (relative extreme deviations), real world datasets often fit into this category; one reason being that outliers are normally considered as unusually large deviations for a given data feature.     

The present work seeks to establish important and practical connections between the approach used by the perception algorithm and prior decades of research in neurophysiology and ANNs; particularly that of information processing in the retina and visual cortex. The idea that the brain performs computations was first put forward by \citet{mcculloch43a} where they proposed that basic neural nets with threshold logic can be combined into complex circuits, just as with logic gates, to compute anything that is capable by a Turing machine. Thus, they introduced the idea that many single units of neurons can combine to give increased computational power. The logical calculus made a number of physical assumptions about biological neurons such as communication via excitatory and inhibitory connections and that certain thresholds must be reached before a neuron itself begins to fire excitatory impulses. The excitations were modelled as all-or-none processes and assumed to occur through more than a single afferent synapse, so that given a sufficient number of impulses arriving at many synapses within a period of latent addition, a neuron can be observed to carry out summation over time between the arrival of the impulses and its own propagated impulse. The inhibition can be absolute or relative, and thresholds fixed or dynamic. Many of their simplifications and assumptions are foundational and commonplace in current ANN research. However, the McCulloch and Pitts (M\&P) neuron model and nets did not provide for a learning algorithm, rather that analysis can realise any network for the computation of a task. Rosenblatt's perceptrons \citep{rosenblatt1958perceptron} were the first to demonstrate that an appropriately designed single layer neural network with non-linear activation functions could learn by experience (supervised learning) to solve perceptual problems using Hebbian learning. Rosenblatt took a probabilistic approach that moved away from the strictly binary inputs of the M\&P neuron, and introduced weighted connections and a simple learning algorithm with feedback to discriminate linearly separable classes. It was later proposed to build stacks of such elementary units in multiple layers to learn non-linear decision boundaries; however efficient learning algorithms were as yet unknown in the early development of neural networks. Modern day neural networks are largely a product of the development of the M\&P neuron and perceptrons, whether they be supervised or unsupervised. 

The ubiquitousness of anomaly detection has led to my position that anomalies are what perception is geared towards due to most information of importance and relevance being contained in them---regardless of the sensing modality. Under our discussion of unusual deviations from randomness, this is when a notion of expectation is formed by a perceptual grouping, and one or more observations `shatters' that expectation by taking on unexpected values with respect to the grouping. The concept of anomaly detection as a human endeavour can be taken a step further when we conceptualise the perception algorithm as a unit of computation akin to the simplified modelling of biological neurons like the M\&P neuron and perceptrons (see Figure \ref{fig:single_neuron} for an illustration). The proposed elementary neuron is essentially an unsupervised learning processing unit that decides which, if any, of its input are anomalous. A key differentiation from prior models being that learning is without supervision as in perceptrons, and without manual or fixed specification of a network for a desired computation as in the M\&P models. The output is also not a single response, but individual to each input stimulus. The neuron is adaptable to the stimuli with data dependent thresholds and every output is a non-linear function of all or a fraction of its inputs. The outputs can be binary (all or nothing) but with each having an associated score that can be both positive or negative; each considered either excitatory or inhibitory respectively. The computations are `intelligent', and not simply a relay of information to some obscure higher processes (such as the human brain). 

The translation into a fundamental computational neuron enables intuitive layering and stacking of multiple such units together to form an unsupervised artificial neural network (see Figure \ref{fig:neural-network} for an illustration).  The inspiration for the neural network comes from descriptions and properties observed about biological neurons, the human retina and cortical structures; and is designed for the detection of meaningful events. The neural network assumes a binary input stream arriving over a unit interval of time at each receptor node, where the streams are summed to integers to be mapped to a decision on whether it is meaningful or equivalently considered to be an anomaly. The neural processing is done first by taking random variable subsamples of the receptor outputs by a sufficiently large number of neurons working in parallel. Each build from their experience a measure of normality by performing local anomaly detection to eject anomalies from their subsample, and then re-learning on the subsample data that remains, so that each neuron represents an expert partial model. The computation of an anomaly score and decision is made for each receptor value as in the perception algorithm, and such scores or decisions are summed by each output node so that positive sums indicate an anomaly, otherwise the observation is considered normal. The neural network is uni-directional and unsupervised, only working with the input data and presently no direct feedback from the output nodes or otherwise is utilised. Robustness, redundancy, reliability and accuracy are also maintained or improved by spreading computations across the network. Furthermore, through the use of sufficient numbers of neurons and variable subsampling the network is also \emph{practically} parameter-free. 

One could ask whether other standalone anomaly detection algorithms can be conceptualised as a neuron model, and stacks built together into a network of partial or full models. Indeed, we could represent any multi-dimensional mapping of points in $\mathbb R^n$ to $\mathbb R^n$ as a neural network model similar to that proposed in the present work provided the end responses are binary decisions or at least numeric scores. Aside from the fact that not all algorithms have been developed and tested as such, network models---also known as ensembles---can present difficulties in use. Base algorithms can be relatively slow and thus unsuitable for interactive use or real-time application. Algorithms may also be relatively expensive in terms of performance and storage making them unsuitable for constrained devices. This becomes exacerbated as an ensemble. Another important consideration is that base algorithms may require data dependent parameters to be set by users. Given the assumption that anomaly detection, and the perceptual learning tackled in the present work is unsupervised, it is difficult to decide how such parameters should be set automatically for every data set encountered. Even if the effect of parameter choices can be stubbed by ensemble techniques such as random variable subsampling, it still leaves the problem of deciding parameter ranges, combining scores, and the setting of thresholds for deciding if observations are anomalies. An important remark here related to parameter choice is whether base algorithms that benefit from partial views can perform local anomaly detection to eject anomalies before re-learning, this is an aspect that can significantly improve results. Furthermore, theoretically most algorithms model the normal data and are thus geared to produce better fits with increased amounts of data rather than by partial views. Finally, the amount of data required by each model of the ensemble or network can vary so that some may end up being more expensive than applying the base model to the entire data set. This is particularly problematic for high dimensional or large modern day data sets.  

One algorithm however that has parallels with the neural network approach of this paper is Isolation Forest \cite{Liu2008}. In this model random subsamples of the data (256 points) are taken to build many isolation trees ($\approx 100$) where each tree then scores every point in the data set, and the average taken as an anomaly score. One could consider each tree as a neuron model, and the forest akin to the network shown in Figure \ref{fig:neural-network}. However, although Isolation Forest is practically parameter-free due to the default setting providing largely optimal results, it is relatively slow and ultimately a contamination ratio parameter must be provided to threshold which output scores indicate anomalies. This can be automatically specified to give good results in many cases, but it can also sometimes give unsatisfactory or extremely poor results. The sensitivity to the size of subsamples also requires further exploration as practically it appears to give good results, but it can be a factor in performance as stated in the original paper. We also note that the original isolation forest does not have scope for carrying out local anomaly detection---should it be beneficial for its performance.

To facilitate the exposition of the neuron model and neural network the rest of the paper is laid out as follows: Section \ref{section:prior_art} reviews unsupervised anomaly detection using ANNs. Section \ref{section:neuron_model} introduces the perception algorithm as a neuron model together with the assumptions made and relations to prior models. Section \ref{section:neural_network} describes the proposed neural network and architecture, together with reasonings for the design choices. A number of properties of the neural network are also discussed in relation to assumptions and findings in biological neural processes. Finally, section \ref{section:conclusion} concludes the paper and ends with a discussion of thoughts for future research.

%
\section{A Review of Anomaly Detection using Neural Networks}
\label{section:prior_art}

The neural network approaches to anomaly detection can be broadly categorised into three with some overlap: supervised, semi-supervised and unsupervised. In supervised anomaly detection the assumption is that labels for the normal and anomalous data are available. However, this approach is not usually applicable in the field of anomaly detection due to the common scenario of data being unlabelled and because of the class imbalance problem. Semi-supervised learning approaches are used under the assumption that some portions of labelled data are available; this could be a combination of both normal and anomalous examples but more often than not it is assumed only the labels of the normal class of data are available. Although applicable in some scenarios, in most real world anomaly detection problems it is unlikely that \emph{only} normal labelled data is available and thus the semi-supervised approaches are either abandoned, continued to be used regardless or are modified so as to have some resilience to contamination of the normal data by anomalies. Unsupervised learning algorithms are the most common neural network based methods for anomaly detection due to the near ubiquity of data being unlabelled and because it is difficult, if not impossible, to characterise and label all cases of anomalies; past, present and future. Hence the present work is concerned primarily with unsupervised anomaly detection. 

\begin{figure}
\centering
  \includegraphics[width=1\linewidth]{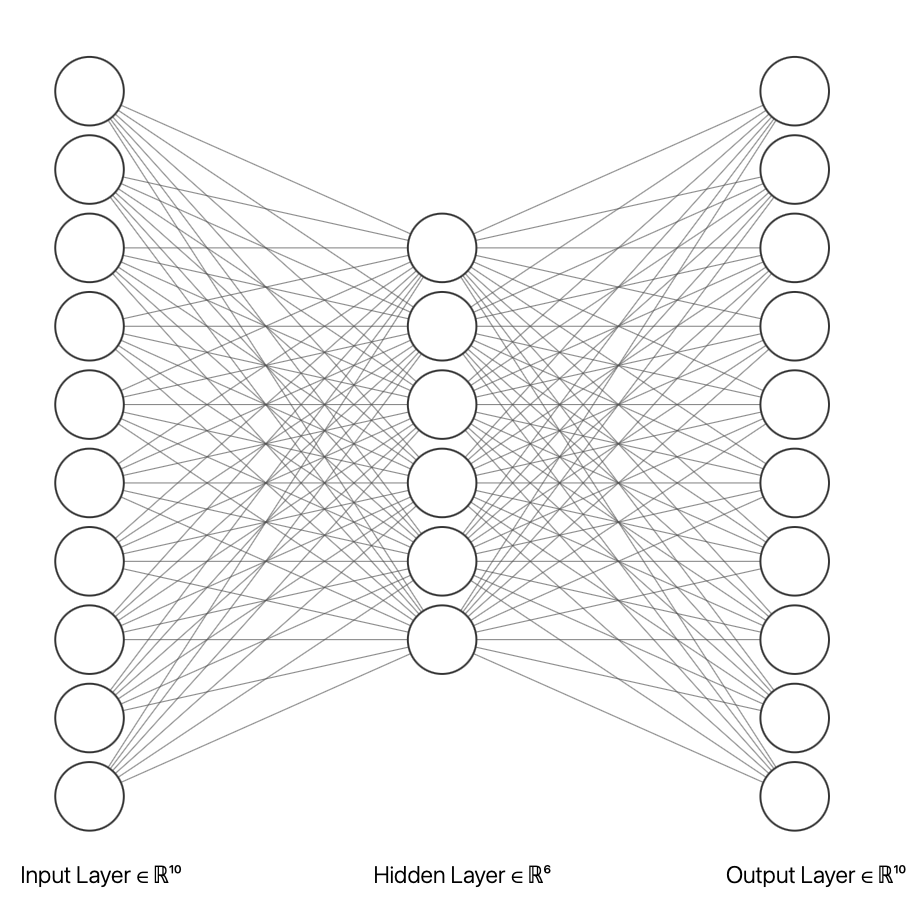}
  \caption{An under-complete autoencoder model with one hidden layer and fully connected nodes between layers.}
  \label{fig:autoencoder}
\end{figure}

Anomaly detection using neural networks are based on autoencoders with the replicator neural network \citep{Hawkins02outlierdetection} being the first to be specifically developed for this application. Autoencoders are unsupervised neural networks that learn a compressed or lower dimensional feature representation space of the input data; a typical example is illustrated by Figure \ref{fig:autoencoder}. An encoder maps the data onto a low dimensional feature space, while the decoder tries to reconstruct the data from this space. The general approach is to design a neural network with the same number of input nodes as output nodes, but with a number of hidden layers. By introducing constraints such as limiting the number of nodes in the hidden layers the network can learn interesting structures. Autoencoders generally set the output values to be the input values so that an approximation to the identity function, $A(x) \approx x$, is learnt using backpropagation and gradient descent on a differentiable cost function that tries to minimise the reconstruction error between the original and reconstructed values. The learning of a non-linear function approximation is accomplished by each hidden neuron summing its input and passing it through a non-linear activation function such as a sigmoid or tanh. The network is ideally trained upon only the normal data or on data that is contaminated with anomalies, but where it is assumed that anomalies are relatively few; as is assumed in anomaly detection problems. The key assumption of autoencoders is that due to the network learning to represent normal data in a compressed representation, anomalies will be reconstructed with larger errors than normal examples from the compressed space. Furthermore, although the data is unlabelled, this method is more accurately described as self-supervised learning due to it using the original input values as the target output values. 

Many variants of autoencoders have been developed using different regularisation approaches. Sparse autoencoders \citep{Ranzato2006} are inspired by observations and theories that only relatively few neurons are active during neural activity and hence they penalise activation units and encourage fewer nodes in the hidden layer to be active. Denoising autoencoders \citep{Vincent_denoising_2008} have been developed to be robust to noisy variations in the data by learning to reconstruct data from corrupted examples instead of the clean original data. This forces the encoder and decoder to implicitly learn the essential aspects of the distribution of the input. Contractive autoencoders \citep{Rifai_contractive_2011} introduce a penalty term that forces the model to learn a function that changes little in response to a slight change in the input data. Applied to the training data this forces the learning algorithm to capture the essential information about the training distribution. Replicator neural networks \citep{Hawkins02outlierdetection} are a specific type of autoencoder utilising techniques from image compression to anomaly detection. The network has three hidden layers with tanh activation functions for the two outer hidden layers, but a staircase like activation for the middle hidden layer that divide continuously distributed data points into a number of discrete valued vectors. This mapping naturally places the data points into a number of clusters enabling outliers to be further analysed by identifying them with their respective groups. Variational autoencoders \citep{Kingma2014AutoEncodingVB} are an interesting theoretical and practical advancement which learn a latent variable model for the input data. These networks learn the parameters of a probability distribution modelling the data to enable the sampling from the distribution to generate new input data samples. In this setup the encoder portion learns two parameters in latent space: mean $\mu$ and variance $\log (\sigma)$. Then random samples of similar points are taken from the latent normal distribution that is assumed to generate the data by $z = \mu + \exp(\log(\sigma)) \times \varepsilon$, where $\varepsilon$ is a random tensor. Finally, a decoder network maps these latent space points back to the original input data. The parameters of the model are trained via two loss functions: a reconstruction loss forcing the decoded samples to match the initial inputs (just like in our previous autoencoders), and the Kullback-Leibler (KL) divergence between the learned latent distribution and the prior distribution, acting as a regularisation term. 

Autoencoders have attracted much interest because they have been considered to have the potential to solve the unsupervised learning problem by learning feature representations via a generic learning algorithm for different sensing modalities such as speech, audio and images. Indeed, the hidden nodes of a multi-layer network can be analysed to show that they appear to be learning successively higher representations of the input data. The method learns automatically from the data without human engineering of features which alleviates this burdensome task; however, this may come at the cost of learning irrelevant aspects of the data, and in the case of anomaly detection may not highlight anomalies of interest.

The use of autoencoders for anomaly detection is subject to a number of theoretical and practical issues. Neural network based algorithms for anomaly detection have been first proposed since at least $2002$ \citep{Hawkins02outlierdetection}, however it is apparent from surveys, evaluation papers and personal correspondence with industrial production teams that such an approach is rarely used in practice \cite{AggarwalOutlierAnalysis2017, Goldstein2016, Chen2017OutlierDW}. A major reason is that neural networks have been traditionally slow to train even on specialised hardware.  Indeed, data analysis is a highly iterative process and developers are likely to favour approaches that yield quick and in many cases near instant results---unless there is a significant increase in accuracy which has not been clearly and universally found to date. However, continuing algorithm and hardware development may make the difference in speed largely inconsequential for all but the most demanding tasks or in environments where resources are severely constrained. Autoencoders are also sensitive to noise and can overfit the training data (particularly on small datasets) where getting stuck in local optima is a real problem \citep{Chen2017OutlierDW}. This reduces their performance when released in production environments. One solution to this problem is to increase the data size, however this may not always be sufficiently possible and an immediate consequence is that training time will increase. Another solution is to use an ensemble of autoencoders with varying random connections between layers \citep{Chen2017OutlierDW}, however this can lead to slower training times and increases the number of parameter choices that need to be made---as discussed next.

The neural network approach to the anomaly detection problem is one of optimisation and requires the setting of a large number of hyper-parameters such as: choice of number of hidden layers; number of nodes in each layer; the optimisation algorithm (ADAM, RMSprop); the activation function types for each layer (linear, non-linear); the choice of activation function (tanh, sigmoid, linear, relu); the learning rate; type of regularisation (dropout, sparsity, number of nodes in a hidden layer); the type of autoencoder (complete, undercomplete, overcomplete, sparse, denoising, variational); number of epochs; batch size; and when to terminate training. The setting of the structure and the bewildering number of parameters is challenging and somewhat considered an art, but one that is largely guided by feedback from the training data to minimise reconstruction loss since autoencoders are self-supervised networks. Another important parameter is the ratio of anomalies expected to be in the dataset or equivalently the threshold at which to consider observations as anomalies. As for most anomaly detection algorithms, this is difficult to specify in advance. The correct setting of thresholds is data dependent and hence highly variable since new data can be expected to contain a relatively large number of anomalies, very few or even none at all. 

Some final remarks on the use of autoencoders is that they were originally designed for compressing data, thus the anomaly detection is a by-product of the original objective and may not be optimised for its primary task. Autoencoders are also highly data specific and hence are applied narrowly only on the data they are trained upon. The data specificity is to such an extent that given a network trained in the domain of images, if for example the training data is human faces, it will not perform satisfactorily on images containing natural scenes. This is mainly an issue where training has been costly and there is need for repurposing models. 

%
\section{The Neuron Model}
\label{section:neuron_model}
%

\begin{figure}
\centering
  \includegraphics[width=1\linewidth]{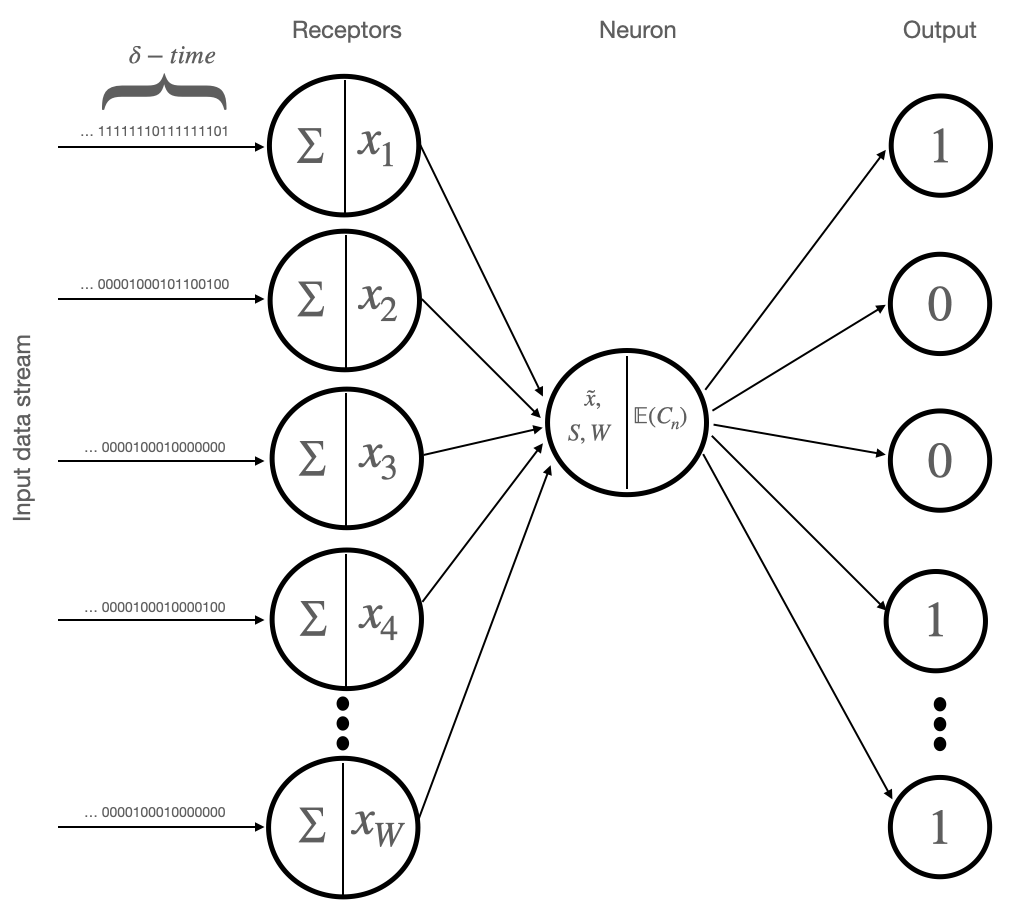}
  \caption{The perception algorithm conceptualised as a single neuron model with an assumed binary input stream of data over $\delta$-time i.e. the input stimuli is assumed to arrive as quanta. Such quanta are summed and output by the receptors. The neuron receives all such outputs to compute the integer median $\tilde{x}$, and yield the parameters $S$ and $W$ after transformation; these are used to compute the expected value for each input. The output here of the neuron is binary, representing whether it deems the spatially corresponding input as anomalous or not.} 
  \label{fig:single_neuron}
\end{figure}

The perception anomaly detection algorithm operates on the principle that unexpected events with respect to a prior expected grouping are anomalies, and it can be conceptualised as a parameter-free unsupervised neuron model that is illustrated by Figure \ref{fig:single_neuron}. In this version the input data is assumed to arrive as a batch composed of a stream of \emph{indicators} (${0,1}$ values) over a fixed period of time $\delta$. Each receptor performs a summation over $\delta$-time to yield integer $x_i$ that is provided as input to the neuron which carries out a simple two stage computation. The first, and considered the fit or learning stage, is to compute the count of the number of observations $W$ which is simply the number of input connections, the integer median $\tilde{x}$, and $S = \sum_{i}^{W}{|x_i - \tilde{x}|}$. This is the neuron's experience of the world over $\delta$ unit time with the median chosen as a robust approximate measure of centrality and parameters learnt without requiring supervision. In the original perception algorithm there is only one time period $\delta$, and all the $x_i$ correspond to the entire dataset. 

The neuron computes for all its input values whether any are considered to be anomalous. In effect every input value has a corresponding output representation on a map that records the anomalous nature of every stimulus; $0$ representing normal instances and $1$ for anomalies. In related models such outputs may be $-1/+1$, or additionally a range of positive and negative $scores$ where increasingly positive scores represent increasingly anomalous inputs corresponding to the meaningful. The anomaly computation is carried out by the neuron under the a-contrario assumption that all such atomic elements composing $S$ are uniformly, randomly and independently distributed amongst $W$ windows; then given the realisation of the input data it computes whether a particular number of atomic elements (an n-tuple) observed in a window is unexpected to have occurred under the a-contrario assumption. Where it is observed that the number of occurrences of the event is not expected to occur even once, yet it has occurred, an anomaly (by definition) is perceived (because it is unlikely to have happened in uniform random noise) by the Helmholtz principle. Specifically what is computed is the expected value of the number of $n$-tuples, $\mathbb{E}(C_n)$, of an event ($n=|x_i - \tilde{x}|$) that occurs. Thus for a given observation $n$, the expected count of $n$-tuples is computed by the following formula where any such $n$ that satisfies it is considered anomalous (see \citet{nassir2021anomaly} for full details):
\begin{equation}
\label{equation:anomaly_detector}
\mathbb{E}(C_n) = {S \choose n} \frac{1}{W^{n-1}} <1
\end{equation}
Equivalently, for computational reasons and after a log transformation we have an anomaly when
 \begin{equation}
 \label{equation:anomaly_detector_log}
f_{S,W}(n) = -\frac{1}{S} \left( log{S\choose n}  - (n-1) log(W) \right) > 0
\end{equation}
The perception algorithm, and hence the neuron model, is general enough to handle any numerical input up to a specified level of decimal accuracy. This is achieved by first transforming all the inputs to integers using rounding and multiplication by an appropriate order of magnitude. Then the algorithm is applied as described for the integer data. 

\begin{figure}
\centering
  \includegraphics[width=1\linewidth]{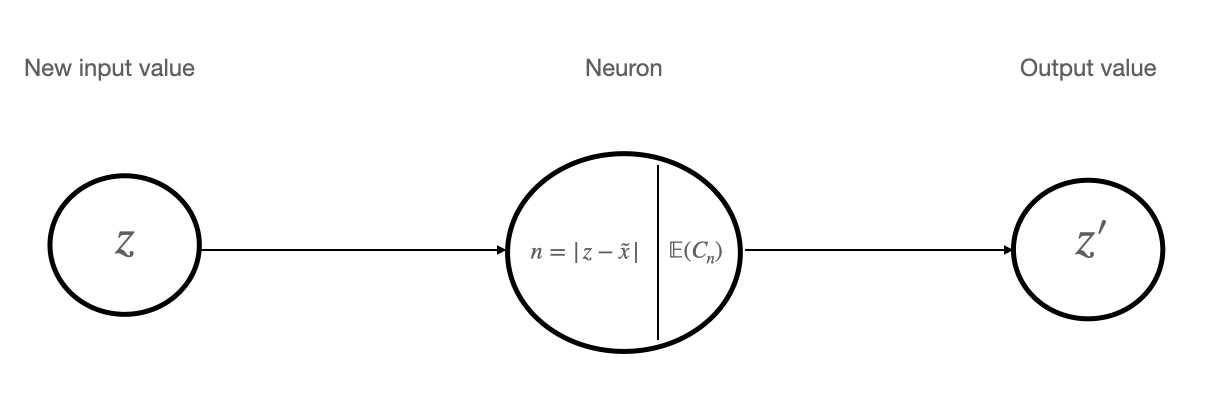}
  \caption{A neuron that has experienced and gained its parameters can also make individual predictions on newly arriving input values. An input $z$ is transformed using the stored median, and the expected value of the $n$-tuple computed using the stored parameters $S$ and $W$ to yield a binary output decision.}
  \label{fig:single_neuron_prediction}
\end{figure}

The decisions made for every input are mapped to the output nodes, however the neuron can also make decisions on new inputs that did not form part of the original experience. This is illustrated by Figure \ref{fig:single_neuron_prediction} where given a new input $z$, it is transformed to $n=|z - \tilde{x}|$ using the median from the learning stage and the expected value of the $n$-tuple is computed using the parameters $S$ and $W$. The output is again binary but also has an associated score indicating the relative extremeness of the observation for scores greater than $0$. 

The neuron model requires some handling and interpretation over different ranges of its input. Where $n=|z - \tilde{x}| \leq S$, equations (\ref{equation:anomaly_detector}) and (\ref{equation:anomaly_detector_log}) are well defined; however for $n>S$ the binomial coefficients are either taken to be undefined or subject to interpretation. Under equation (\ref{equation:anomaly_detector}) the binomial coefficient can be taken to be $0$ with the natural interpretation that there is no way to choose $n$ items from $S$ given that $n>S$, and leads to $\mathbb{E}(C_n)=0$, giving us a constant anomaly score and always an anomalous decision. Hence we lose information about increasingly anomalous inputs. However, following equation (\ref{equation:anomaly_detector_log}) which is what is used in practice to compute the expected scores, the log of the binomial coefficient is undefined at $0$; for practical reasons $log{S\choose n}$ is taken to be $0$ so that the computation will be $>0$ and the function linear. Although we lose the original relation between anomalous scores of the input with increasing $n$, it does preserve the increasing anomalousness. Practically the anomaly \emph{decisions} remain the same for $n>S$, only the interpretation of the scores differ compared to the range $0 \leq n \leq S$. Furthermore, this is all of concern where newly arriving data is processed without learning. Where the computations are carried out afresh there will be no binomial coefficient issue to handle as it will be defined for all its inputs.

\begin{figure}
\centering
  \includegraphics[width=1\linewidth]{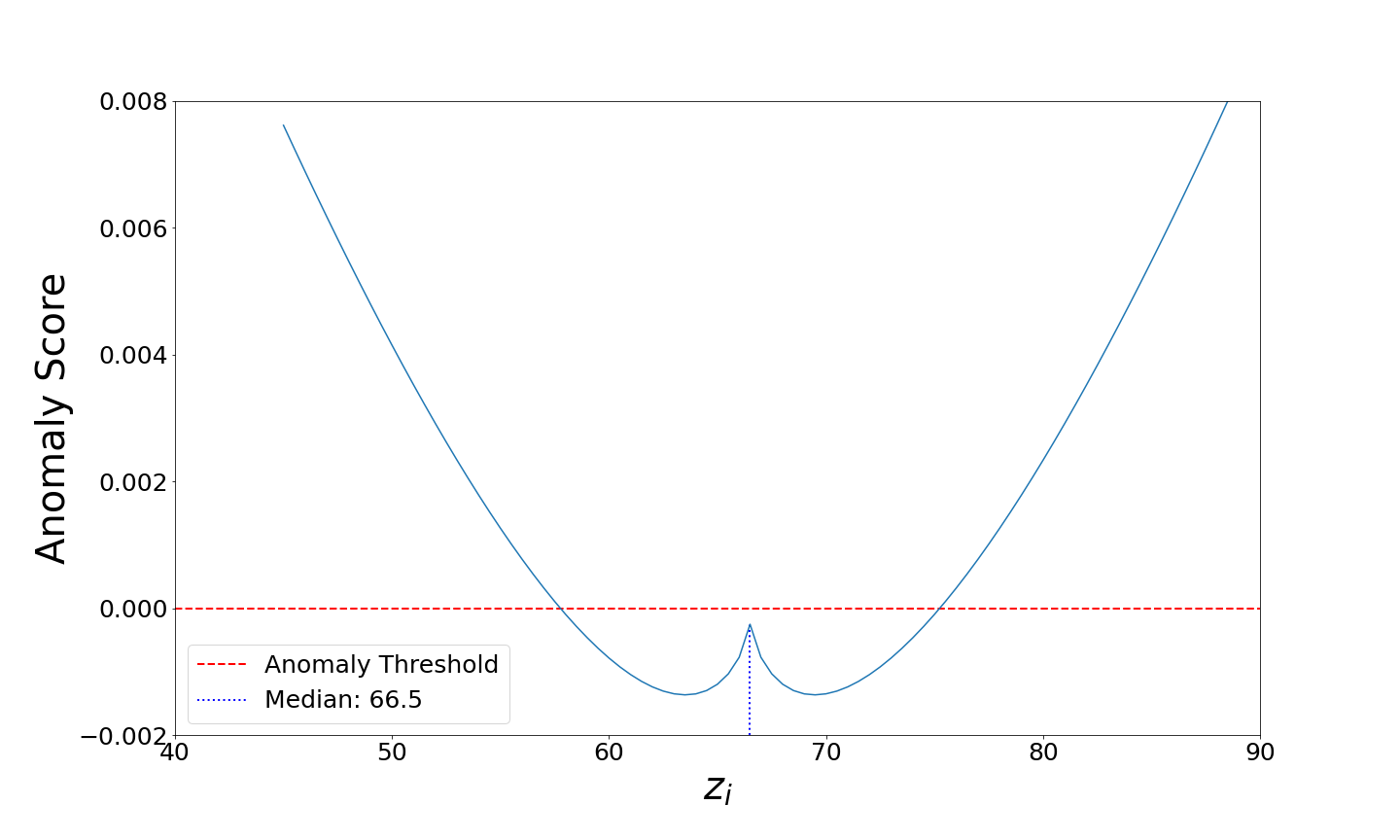}
  \caption{Graph of the neuron function over a portion of its defined input.}
  \label{fig:neuron_graph}
\end{figure}

\begin{figure}
\centering
  \includegraphics[width=1\linewidth]{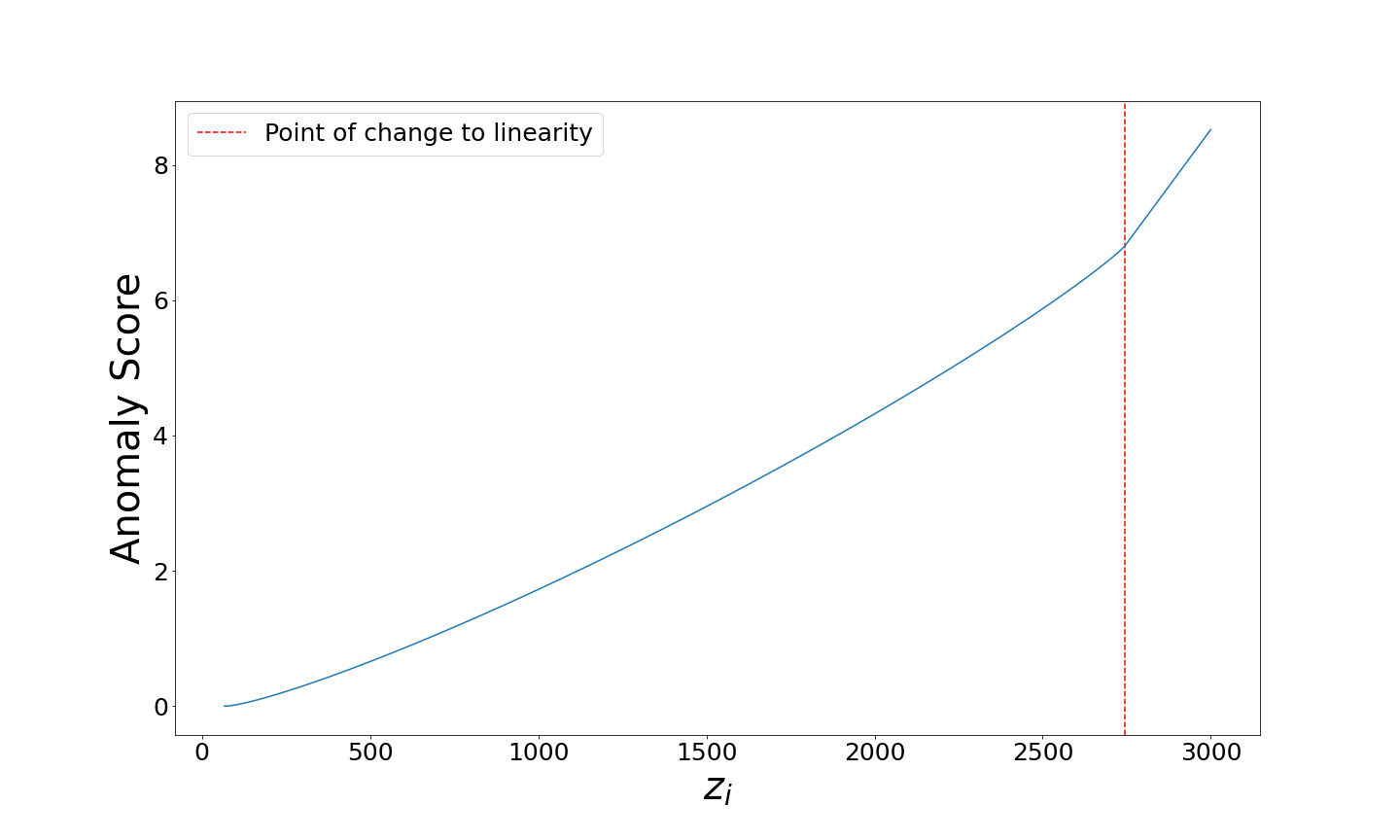}
  \caption{The right portion of the graph of the neuron function including where it artificially turns linear at the point where $n=|z_i - \tilde{x}|>S$, indicated by the vertical dashed line.}
  \label{fig:neuron_graph2}
\end{figure}

Figure \ref{fig:neuron_graph} illustrates a typical example of the function $f_{S,W}$ over a range where $n=|z_i - \tilde{x}|<S$. The function graph is symmetric around the median $66.5$ (which is transformed to integer $665$ by the neuron), and although the median value is below the anomaly threshold it is not the most normal input since it is only used to approximate the data centrality. Indeed the anomaly score dips further before accelerating higher with increasing or decreasing $z_i$; note the curvature of the graph. Once past the anomaly threshold score of $0$, any input value is considered anomalous with an increasing associated score. Figure \ref{fig:neuron_graph2} illustrates a typical example of $f_{S,W}$ over a much larger range of input including where $n=|z_i - \tilde{x}|>S$ and the function changes to linear. Only the right portion of the symmetric graph is shown. As previously noted, the function is in fact undefined over the range $n>S$, but artificially modified for practical purposes to predict scores over newly arriving input. 

It is interesting to note the similarities between the neurons non-linear function over range $n \geq 0$ (including the linear component), and that of the concept of activation functions in ANNs. Activation functions, sometimes described as squashing functions, are an essential component of ANNs to ensure the model can represent non-linear relationships in the data that a linear classifier can then handle. Examples such as the rectified linear unit (relu), leaky relu, parametric relu, exponential linear units, swish, gaussian error linear unit and the scaled exponential linear unit are of interest due to an essential similarity in shape, and it is intriguing to reflect on the neuron model computation as one of activation in unsupervised and supervised learning. 

\subsection{Relation to Prior Models}
The inspiration for the neuron modelling comes from neurophysiology \citep{Hubel1959ReceptiveFO, Barlow72}, computational modelling \citep{mcculloch43a, rosenblatt1958perceptron} and computer vision \citep{Mar82} studies of information processing in the early stages of vision found in the retina and visual cortex. Indeed, the neuron model has many interesting properties that have biological counterparts which will also be elaborated upon in section \ref{section:relations_to_bio_nets} under the context of a neural network. However, for completeness the principle assumptions and properties that borrow, add and differ from the prior M\&P neuron and perceptron models are given here: 

\begin{enumerate}
\item \emph{Dual activity}. The present work assumes the essential firing of (at least the relevant subset of) biological neurons to have both continuous and discrete facets, and that individual biological synaptic endings are specialised so that some are excitatory endings while others inhibitory. The neuron model excitatory output is selective towards the anomalous observations such that it only transmits excitation to a relatively few output nodes and to the rest it transmits inhibition. The activity of the neuron model is foremost an \emph{all or none process} as with the M\&P neuron and perceptrons, but also has a related energy output or score. Thus, the higher the score for a particular observation---beyond the anomaly threshold---the more anomalous it is considered to be.  

\item \emph{Discrete time intervals}. McCulloch and Pitts assumed that the only significant delay in neural activity was synaptic delay and that computations are carried out in discrete time intervals. This discretisation serves our purposes of counting arriving stimuli over a fixed length of time (window length), enabling the probability interpretation and derivation of the perception algorithm, and hence the neuron model. All computations are carried out in discrete time intervals in tune with the assumed discrete nature of the input stimuli.

\item \emph{Immediate computation}. It is assumed in this neuron model that the function is only dependent upon that which is immediately received within a discrete period of time. Thus there are no previous inputs that directly stimulate the neuron. This is identical to the M\&P neuron, but does differ from some perceptron models that considered longer range connections. The idea of immediate computation is also related to discoveries pointing to a large part of the sensory machinery residing in the retina rather than in complex higher processes \citep{LettvinFrog1959}. Thus the neuron itself is designed as to perform essential discriminatory computations based on its immediate input and directly from its sensory stimuli. 

\item \emph{Number of afferent synapses}. The M\&P neuron assumed that a certain fixed number of synapses must be excited within a period of latent addition in order for excitation of the neuron to occur, and that no case is known by only a single synapse. This has an interesting correspondence with the neuron model since one or two input values cannot cause the neuron to fire, but at least three input values are required to illicit a response; and only if one is meaningfully different to the others. The neuron is capable of firing both when there are limited numbers of input connections, and when there are large numbers.

\item \emph{Spatial summation}. Computational models of biological neurons assume a summation is carried out over its inputs. Indeed, spatial summation models have been proposed where impulses may arrive at different points of the cell body or dendrites, and these are summed to trigger an impulse. Summation (after a simple transformation of centering and taking the magnitude) is also an essential component of the neuron models learning stage (but where contrast is required for firing). Note that a record of the stimuli is not required as only the sum is utilised together with the median and count of the number of afferent synapses. 

\item \emph{Adaptive thresholds}. Biological neurons are assumed to have thresholds that need to be exceeded through afferent excitatory impulse summation to yield firing. Similarly the neuron model has thresholds but which are data dependent (adaptive), inherently dynamic and selective to the output. Indeed, if there is insufficient contrast in the stimuli then the neuron will fail to fire regardless of the intensity. Furthermore, the selection of the threshold is arrived at completely through unsupervised means. Hence, the threshold does not have to be specified by problem analysis (as in the M\&P model), nor is it learnt through supervision using large numbers of labelled examples (as in perceptrons). 

\item \emph{Non-linear activation}. It is assumed that the biological nerve impulse is a non-linear response to stimulation, and this non-linear property has been an essential component in the design of activation functions for ANNs. The neuron model response to stimulation is no different and is a non-linear function akin to a ``squashing function". This is closer to modern neural network activation functions than the step threshold functions used in the M\&P models and perceptrons.

\item \emph{Sparsity}. A widely held property of biological neural network activity is that the response of the network to stimuli is sparse i.e. only relatively few neurons are excited. Although, this is not used directly to arrive at the neuron model, it is nonetheless an inherent property where firing occurs only where there are unexpected observations that differ sufficiently from the majority of the input. Thus, either the neuron model does not fire due to too much uniformity of signal or noise in the input, or it fires correspondingly to only those contrastive and rarely occurring observations. Hence when active, only a sparse number of output nodes are excited.

\item \emph{Reliability}. The neuron model is a reliable unit that does not possess any internal random element that may introduce noise into its computation. Any noise originates externally, and its expected value computations are entirely determined by the distribution of the input stimuli.  

\end{enumerate}

%
\section{A Neural Network Model}
\label{section:neural_network}
%

Neurophysiological investigations over many decades have discovered a number of intriguing, surprising and important processes and properties of visual systems. In particular, the organisation and neuronal processing found in the retina and visual cortex have provided inspiration for many models in computer vision and machine learning. The present work builds upon this through the design of unsupervised artificial neural networks made for detecting meaningful observations---or equivalently anomalies corresponding to objects of interest. The (first of many) neural network model presented in this paper assumes that the input data is numerical, has no particular ordering and that normal points form a single grouping; in essence the detection is of global point anomalies represented as unexpectedly deviant values. The network is built from stacking together many elementary neurons (described in section \ref{section:neuron_model}) with the exact architecture guided by research results in neurophysiology, neural networks, ensemble learning and exploratory analysis of the behaviour of the network. The underlying goals of developing such networks are to achieve higher detection accuracy (as measured by a suitable metric for unsupervised anomaly detection), to have more robustness to data extremities, to compute in simple parallel processes that are more natural and easier to modify, to provide redundancy to failure of sufficiently low numbers of parallel components, to provide graceful global degradation by failure of any components, to demonstrate only random variable subsamples of the input data is required for anomaly detection and only a roughly constant amount as data size increases, to model some aspects and assumptions of biological neural processes, and to show that anomaly detection can be carried out by quasi-independent neurons that each compute simple operations and gather little pieces of evidence from small experiments to be combined by subsequent nodes. 

The present section will describe the chosen architecture, detail the reasonings behind the design decisions, and relate it all to hypothetical and factual neurophysiological and computational properties of the visual and nervous systems. 

\subsection{The Model Architecture}
\begin{figure}
\centering
  \includegraphics[width=1\linewidth]{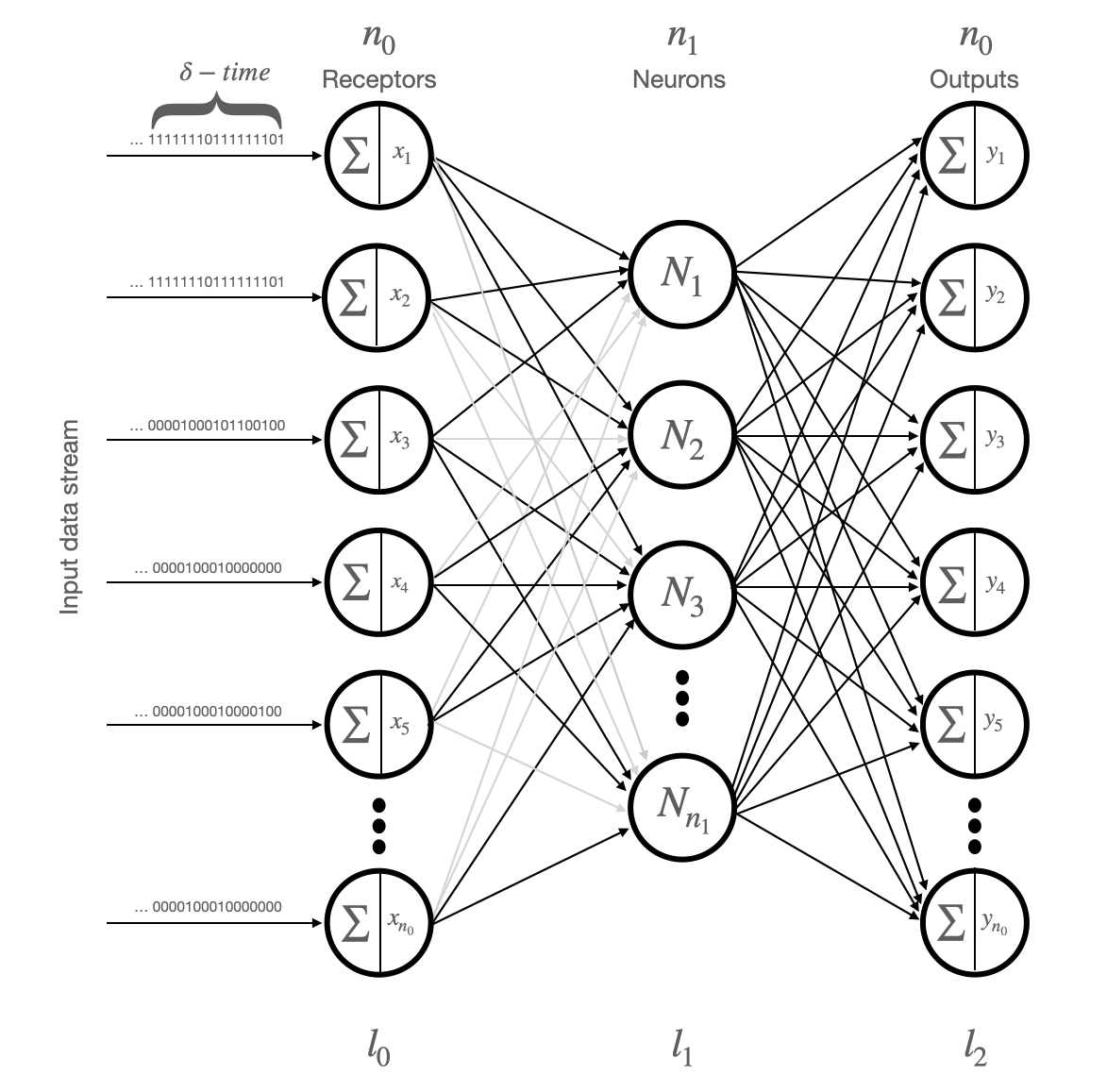}
  \caption{An unsupervised neural network for the detection of anomalies. The one-dimensional feature data ($x_1, x_2, ..., x_{n_0} $) is assumed to be composed of indicators arriving with a uniform distribution over unit time $\delta$, this is analogous to photons arriving at an array of photoreceptors then summed over unit time. The black lines between layers $l_0$ and $l_1$ represent the random input subsample from which a particular neuron learns and predicts upon, while the grey lines indicate inputs that are received by the neuron \emph{only} for making predictions upon.}
  \label{fig:neural-network}
\end{figure}

The network architecture of an unsupervised neural network for univariate input data (such as individual feature vectors) is illustrated by Figure \ref{fig:neural-network}. This is a feed forward network with no feedback learning from its outputs; instead unexpected variations in the data are detected directly by feeding the input data through random variable subsampling and nonlinear processes to immediately arrive at a ranking or a decision on if each observation is anomalous or not. The learning is carried out by each neuron learning three simple parameters; the sum of the subsampled \emph{transformed} portion of values it receives (the total `energy'), the median of these values and the number of corresponding connections. This seemingly little information is enough to learn a model of the world that each experiences and make predictions on \emph{all} and \emph{new} stimuli.  

The input layer $l_0$ composed of $n_0$ \emph{receptors} each sum a stream of assumed uniformly distributed indicators over a fixed window of time $\delta$ and thus hold an integer value of counts $x_i$. Any univariate data vector presented to the network is essentially assumed to have been generated by such stimuli arriving over a unit window of time. The biological inspiration and perspective here is of photoreceptors (being sensitive to even single quanta on average) summing the number of photons arriving over a discrete unit of time, which are then converted into intensity values of nonnegative numbers corresponding to the amount of energy received.

The second layer $l_1$ is the key processing layer composed of $n_1$ neurons (as in section \ref{section:neuron_model}), that each connect fully to $l_0$ but only learn on a randomly selected variable number of receptor values. Generally $n_1$ is much smaller than $n_0$; empirically it is found that a small number of neurons ($\approx 256$) are enough for good results. The sizes of the random variable subsamples is taken from empirical findings to be in the region of $10$ to $1000$ but with a log-normal distribution bias towards smaller values. Each neuron considers each receptor value to be composed of counts of its constituent elements and its processing is hence to determine if the value is unexpected or not and to compute an associated score that is mapped to the output nodes. The neuron is thus connected to $n_0$ output nodes in layer $l_2$. 

The neuron computation is carried out in two phases. Firstly, it performs anomaly detection only on its subsampled portion of data to eject points considered to be anomalies and hence which may hinder learning. Then upon the data that remains it learns its parameters after which it computes for all input values. Note here that the neuron computation has two facets; a score where anomalies have $>0$ and normal points $\leq 0$, and a translation of these to binary values of $\{1,0\}$, or in this particular network it is taken to be $\{-1,1\}$. The former effectively considers neuron firing output to be continuous or composed of a number of quanta that can negatively or positively affect, while the latter assumes the commonly assumed simplification that neurons fire in an all or none (or inhibitory) fashion. 

In the present network each output node sums all the receiving scores from its connecting neurons. Positive values contribute to the excitation of the node while negative values contribute to the inhibition of the node. The total sum is used as a score to rank all the receptor values, with greater scores being more anomalous. Summation, as opposed to taking the maximum or median is chosen due to the belief that an aspect of neural computation is the summation of signals, and that if this exceeds a threshold then firing occurs. Furthermore, taking the maximum leaves the node too sensitive to even a single rogue score while taking the median perhaps leads to a loss of information. Empirical results using summation also provided increased support for taking this approach. 
In order to arrive at an anomaly decision, the sign of the total sum can be used such that if it is positive then the node produces a value of $y_i=1$ to predict an anomaly, otherwise $y_i=0$ to predict a normal observation. However, for slightly increased stability the binary $\{-1,1\}$ computations received by each output node from the neurons are summed instead of the scores in the present network, so that a positive sum decides to class the observation as an anomaly, otherwise normal. 

An important remark on the neural network shown in Figure \ref{fig:neural-network} is that it is easily extended to multivariate data where now each neuron is connected to each feature value of every observation, but again learns only from a subsample. The computation is largely identical to the multivariate version of the perception algorithm detailed by \citet{nassir2021anomaly}, except that subsample anomalies are detected and ejected before re-learning. As before, each neuron produces a score and binary decision output for every multi-dimensional input observation. 
	
It is interesting to compare this neural network with that of the general structure of autoencoders (the other major type of neural network used for anomaly detection) reviewed in section \ref{section:prior_art} and illustrated by Figure \ref{fig:autoencoder}. While both may look similar the functional aspects are very different. The learning in autoencoders is typically carried out over many examples of the data and not on an individual feature, with the output set to be the same as the input, and weights between nodes updated by gradient descent on an optimisation function where the gradients are obtained by backpropagation; thus making the network learning function bidirectional. Many hyper-parameters of the network have to be set with the fundamental goal of obtaining a compressed representation of the input data by the middle layer. In particular a threshold is required to decide when something is anomalous which the network does not innately provide. Although such networks work without labelled data they are better described as self-supervised networks rather than unsupervised. In contrast, the neural network model discovers anomalies over entire individual feature vectors where the data is univariate (or transformed vectors where data is multi-dimensional), with no weights between nodes or feedback learning from labels. Rather than an anomaly being that which cannot be reconstructed well by the compression layer of the network, it is taken to be that which is unexpected to occur under the a-contrario model of uniform random noise. The unsupervised learning is performed over the input data only and predictive scores and decisions sent to the output nodes in an essentially parameter-free and uni-directional method. 

\subsection{Reasonings for the Neural Network}
One of the main contributions of the present work is that of conceptualising the perception algorithm as an elementary neuron and the stacking and organisation of such computational units for the purposes of anomaly detection. Many issues can be raised in the design. One relates to whether introducing additional neurons in the middle layer (with reference to Figure \ref{fig:single_neuron}) can lead to significant performance improvements or desirable properties over the single neuron model. The naive method of duplicating the neurons provides important redundancy against failure, but the exact replication of inputs, computation and outputs gives no improvement in detection. However, the following examples will illustrate how and why the addition of neurons can be beneficial provided all the neuron receptive fields are `diameter limited' in the sense of random fractions of the input space, and hence learning from different portions of the data. 

\begin{figure}
\centering
  \includegraphics[width=1\linewidth]{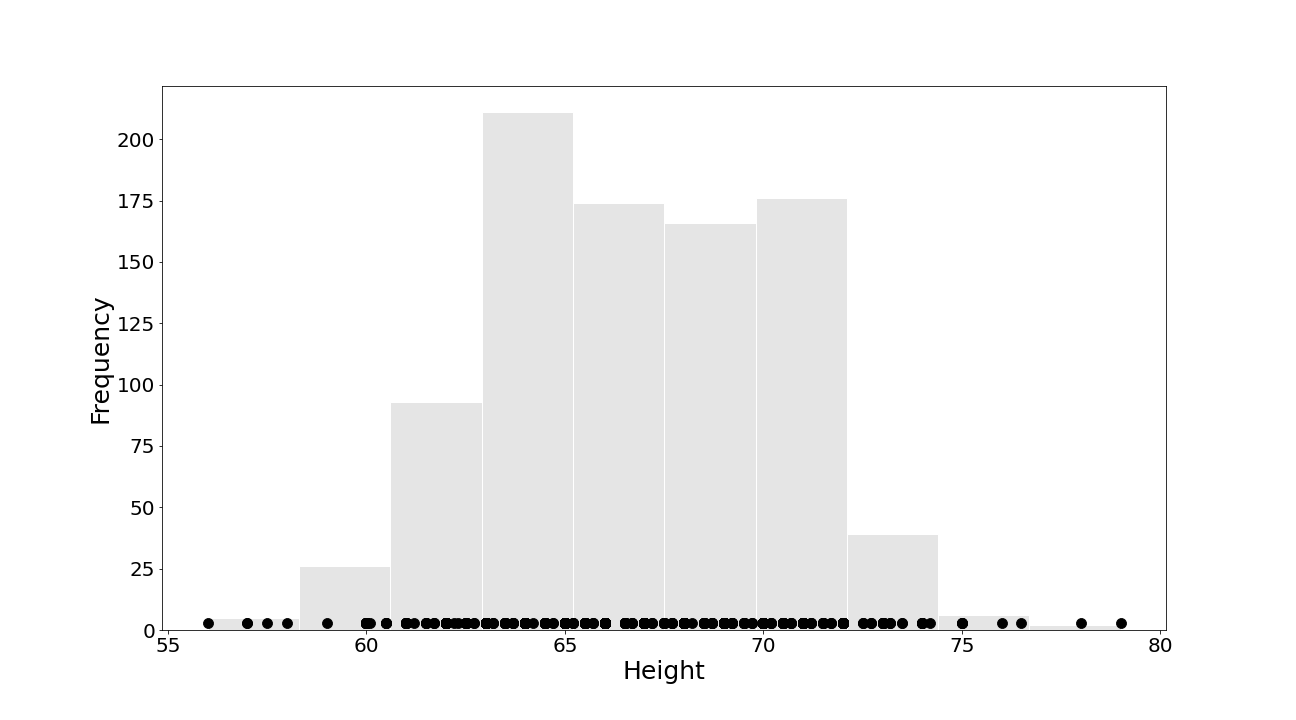}
  \caption{The Galton height data set shown as a scatter plot and histogram. The observations at the fringes of either end of the distribution are considered candidates for being anomalies. The data is unlabelled.}
  \label{fig:galton_heights}
\end{figure}

To convey how accuracy and robustness can be improved, consider first the unlabelled Galton height dataset composed of $898$ observations whose distribution is illustrated by Figure \ref{fig:galton_heights}. The single neuron model is capable of producing good results (decided subjectively) on the dataset to return the anomalous set of observations $\{56,  57,  57.5, 76, 76.5, 78,  79\}$. However, the introduction of a single extreme anomaly $\{700\}$ can have unintended consequences on the neuron output where it now only finds the following anomalies: $\{78, 79, 700\}$. The extreme anomaly has in effect enlarged the decision boundary and hence its presence has masked the other anomalies. This can also be understood from the perspective of precision and recall, where the former can increase at the cost of the latter, assuming observations get increasingly anomalous moving away from the central mass of points. While such an event may appear contrived and could be identified by a domain expert exploring the data for removal or accommodation, the automated analysis and detection of anomalies is required where data streams are too numerous, large or complicated to be all handled by human experts. Furthermore, the extremeness of anomalies need not be so drastic as to make their identification clear and obvious using rule sets. For example, two extreme anomalies of $\{200, 250\}$ can alter the single neuron model results---all be it with less effect---to yield the anomalous set $\{56, 76.5, 78,  79, 200, 250\}$ 
 
\begin{figure}
\centering
  \includegraphics[width=1\linewidth]{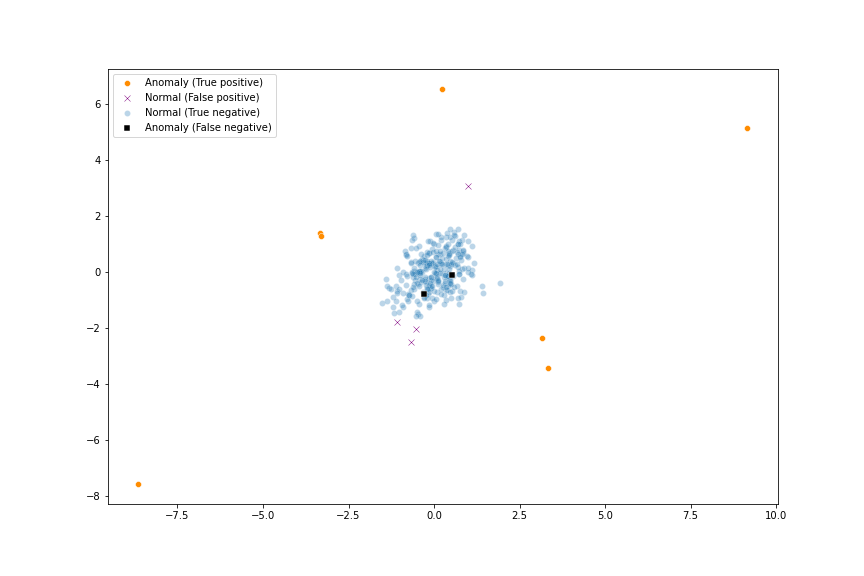}
  \caption{The behaviour of 307 servers measuring the throughput (mb/s) and latency (ms) of response from data set ex8data1 \cite{NgML}. A central mass of points is observed with anomalies being generally distant from the cluster. Overlaid are the results of the single neuron model which achieves an AUC score of $0.93$. The neural network model achieves only $0.88$ on average (keeping in mind the relatively small size of the data).}
  \label{fig:ex8data1}
\end{figure}

\begin{figure}
\centering
  \includegraphics[width=1\linewidth]{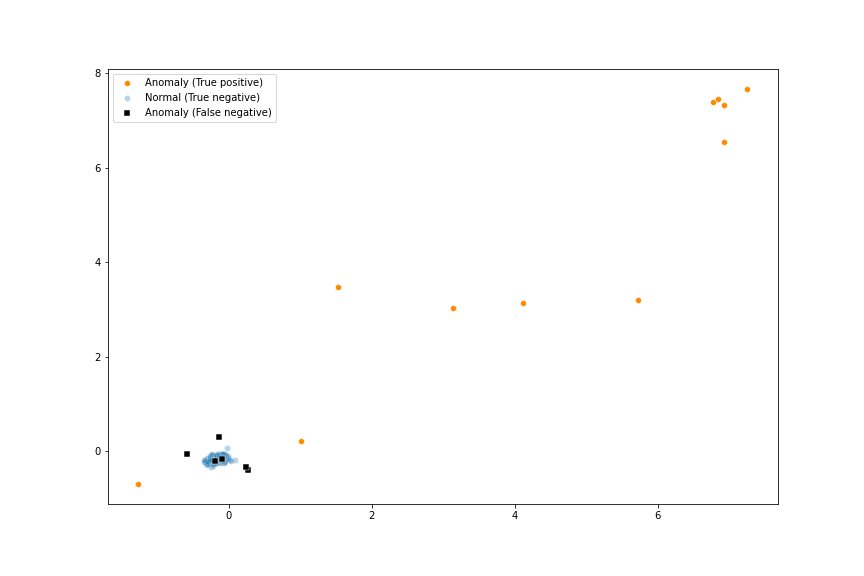}
  \caption{A plot of dataset ex8data1 \cite{NgML} with additional extreme anomalies. Overlaid are the single neuron model results. Not only does the visualisation perspective change but also the anomaly scores of the observations such that the single neuron model achieves an AUC score of $0.71$ and detects only the more extreme anomalies that were introduced. The neural network model by contrast achieves an AUC score of $0.94$.   }
  \label{fig:ex8data1_extreme}
\end{figure}

A second example to illustrate the effect of extreme anomalies is ex8data1 \cite{NgML} that records the labelled behaviour of $307$ servers measuring the throughput (mb/s) and latency (ms) of response. The distribution is illustrated by Figure \ref{fig:ex8data1} and overlaid are the results of the single neuron model. Considering only the Area Under the ROC Curve (AUC) metric, the single neuron model achieves an excellent score of $0.93$. However, with the introduction of some extreme anomalies the visualisation of the data changes (see Figure \ref{fig:ex8data1_extreme}), and the single neuron model now achieves an AUC score of $0.71$. Note the perfect precision but decrease in recall. 

\begin{figure}
\centering
  \includegraphics[width=1\linewidth]{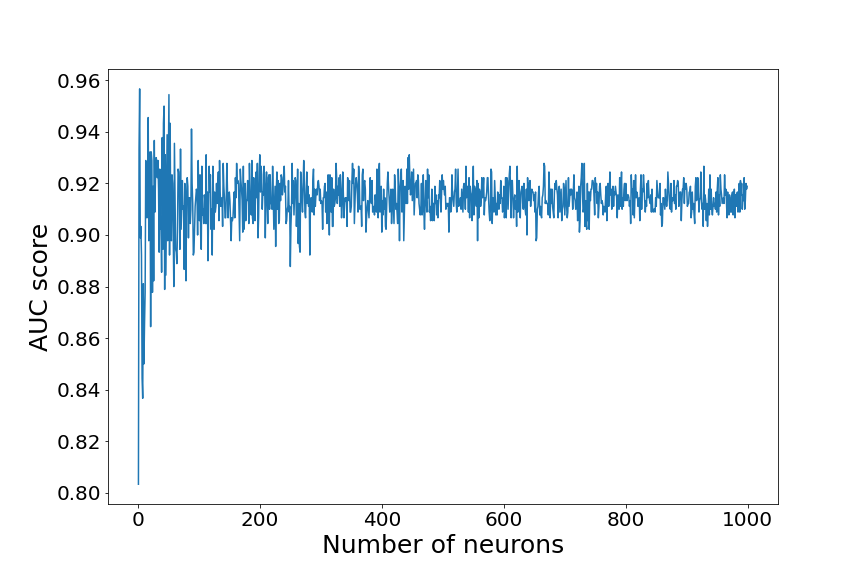}
  \caption{A typical example of the reduction in variance of the AUC score as the number of neurons in the network increase. 
  }
  \label{fig:neuron_var_reduction}
\end{figure}

\begin{figure}
\centering
  \includegraphics[width=1\linewidth]{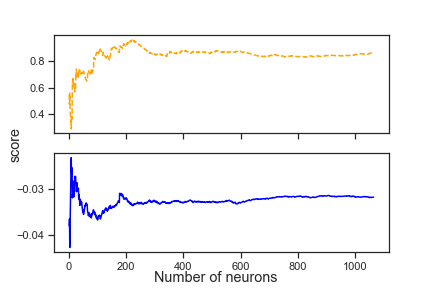}
  \caption{A typical example of the reduction in variance of the network output score for a single anomaly (top graph) and normal example (bottom graph). Note that predicted anomalies and normal examples are given positive scores and negative scores, respectively.}
  \label{fig:single_point_var_reduction}
\end{figure}

The proposed solution to increase robustness of the network and hence accuracy is composed of three parts: (1) the introduction of additional neurons that each perform the same computations, (2) that each neuron learns from a random variable subsample of the input data after ejecting anomalies, yet produces scores and binary $\{-1,1\}$ predictions for all inputs, and (3) that subsequent nodes in layer 2 sum the output scores and binary predictions to enable anomaly ranking and decisions for all inputs, respectively. Neural network type architectures such as this are important for unsupervised learning because of the limited availability of ground truth to update parameters that better reflect beliefs about the environment. Large numbers of partial models provide stability and accuracy likened to supervised learning methods where massive amounts of labelled data are used to guide function approximation. Taking an ensemble of neurons approach we can obtain a reduction in the variance of the sums computed by the output nodes in the neural network. An illustration of typical variance reduction as the number of neurons increases is shown by Figure \ref{fig:neuron_var_reduction} where the AUC score settles towards a specific value or range. Empirically it has been found that $256$ neurons is  enough to provide stable, reliable and accurate results. In addition Figure \ref{fig:single_point_var_reduction} shows the variance reduction of the network output score for a single anomaly and normal example as the number of neurons in the network increases.

\begin{figure}
\centering
  \includegraphics[width=1\linewidth]{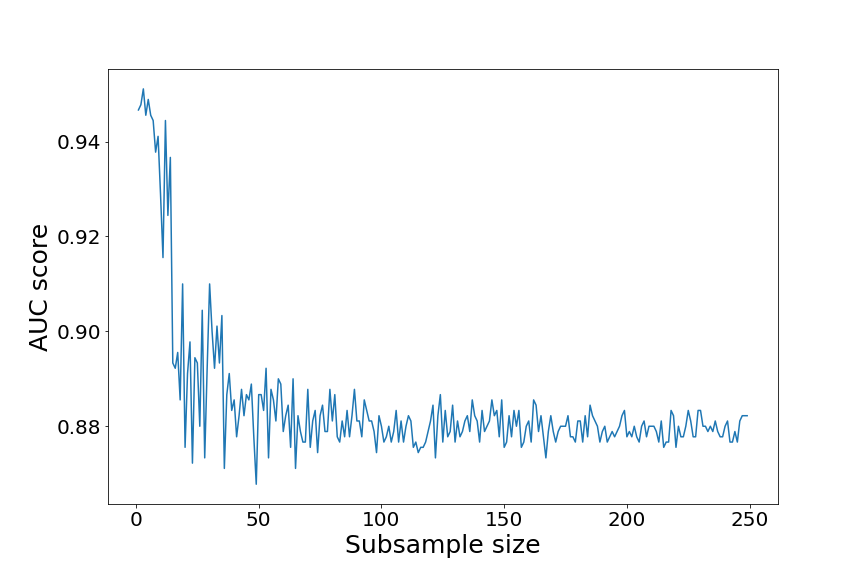}
    \includegraphics[width=1\linewidth]{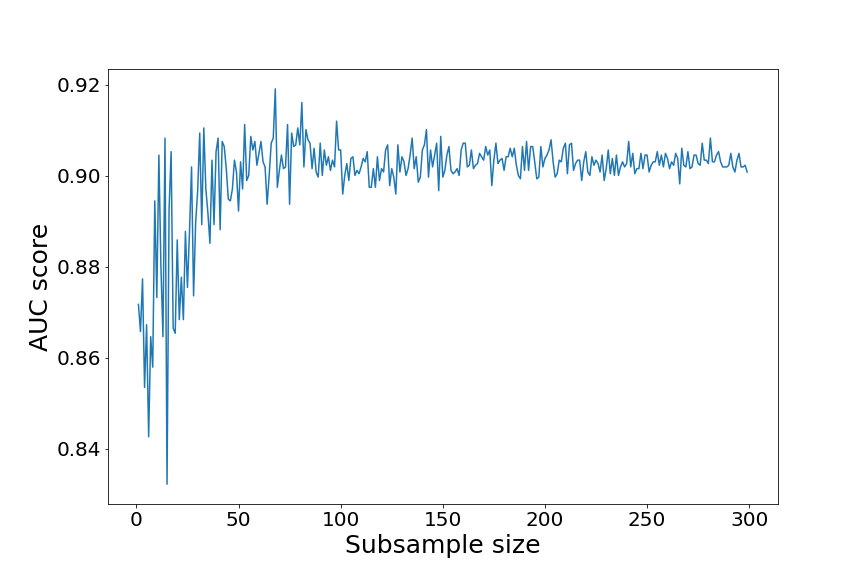}
  \caption{Plots of AUC score over fixed subsample sizes of the neural network for two different data sets. Note that performance differs at different subsample sizes but approaches that of the single neuron model as the subsample size increases. Furthermore, the correlation between AUC score and subsample size can be seen to differ between different data sets.}
  \label{fig:auc_over_subsample}
\end{figure}

The random variable subsampling component of the model is also important to provide increased robustness and hence accuracy to the anomaly detection. This is because the single neuron model (the perception algorithm) can be adversely affected by extreme anomalies that can effectively enlarge the decision boundary (points inside considered normal, and those outside anomalous) and thus lead to increased false negatives; often described as the masking phenomenon in outlier detection. Even without extreme anomalies the diversity introduced by different neurons subsampling different portions of the data can lead to better results. Indeed, variable subsampling is chosen rather than a constant subsample size for each neuron due to the bias effect that sample size has on anomaly detection. The AUC scores can vary considerably over the entire subsampling range for different data sets, where near constant performance is achieved for larger subsamples that approach or match the single neuron model. Examples of the graphs of AUC scores over different \emph{fixed} subsample sizes is shown in Figure \ref{fig:auc_over_subsample} for different datasets. Such findings motivated the use of variable subsample sizes in the region of $\{min(10, N),min(1000, N)\}$ (where N is the length of the dataset) but over a log-normal distribution ($\mu=3, \sigma=2$) to give more probability to lower subsample sizes. The random variable subsampling with this distribution bias reduces the masking problem because each neuron learns from different portions of data and since anomalies are rare the majority of subsamples will be populated mostly by the normal data and without the extreme anomalies. Empirically this also provides a better AUC score using partial models and removes the requirement for users to specify a subsample size parameter that is unknown in unsupervised learning problems for optimal performance. Thus the neural network is kept practically parameter-free. 

An ensemble of independent neurons using random variable subsamples of the data to carry out parallel computations of the input stimuli implies that each neuron experiences a likely different view before computing all the independent predictive scores. This design gives two important properties of redundancy and graceful global degradation. The former implies that removal or failure of sufficiently few neurons and connections keeps the network performance practically the same. The latter property implies that degradation of the network performance is the same---on average---with the continuous removal or failure of any neurons or connections, and the network still able to perform reasonably well provided enough remain. Both these properties are illustrated by Figures \ref{fig:neuron_var_reduction} and \ref{fig:single_point_var_reduction} where it can be seen that reducing the network by a modest number of random neurons from $256$ still maintains the network performance and score in a reasonable range. 
An ensemble of neurons working in such a manner thus provides redundancy without exact replication for the simplified modification, repair or failure of the units. Indeed, in addition to handling inaction of a neuron, working in simple parallel processes enables individual neurons to compute more complex and local functions that can be combined for global solutions. Furthermore, the formation of collective experiences and properties of redundancy and global degradation has utility and advantages for distributed processing applications.  

An interesting consequence of the use of random variable subsamples and $\approx 256$ neurons is that as the data size increases only an approximately constant fraction of total samples are required. While this may be relatively large for smaller data sets, it can be a small fraction for larger ones. For example, given a data set of $1000$ examples the total number of samples is in the order of $10^3$ which is considerably more than the size of the data. However, this remains the same for a data size of $100,000$ examples where we have a $10$ fold reduction in data usage. A single neuron model would utilise the entire data set in both cases whereas the neural network can provide better results with considerably less data. This indicates that for anomaly detection not only is all the data unnecessary for learning, but also that connections between input, computational units and output need not be entirely reliable since performance can in fact increase by using random variable subsamples (taken over the entire data set!).  

Coming back to the Galton dataset with extreme anomalies, the results set found by the neural network is $\{56,  57,  57.5, 58, 75, 76, 76.5, 78,  79, 700\}$ and for ex8data1 with more extreme anomalies (Figure \ref{fig:ex8data1_extreme}) the AUC score is $0.94$. Both examples illustrate the robustness of the network against extreme anomalies and the possible improvement in detection and performance over the single neuron model. However, it is important to keep in mind that these are only illustrative examples and that they are not necessarily the ideal data sets to apply the network model to due to their relatively small sizes. Other small data sets experimented with in private indicate a similar trend but in some instances the neural network always gives similar results to the perception algorithm, and for others it already performs better with and without adding extreme anomalies.
%

\subsection{Relations to Biological Neural Networks}
\label{section:relations_to_bio_nets}
The neural network proposed in this paper does not claim to model complex biological neural networks. However, it does take inspiration from such networks and seeks to model certain elementary and simple aspects that may increase its applicability and improve performance. Thus it is worthwhile relating some of the network properties and exploring how these can be beneficial for anomaly detection and those problems that may be desired to be addressed in future. 

\begin{enumerate}
\item \emph{Random, parallel and independent processing:}
It has been observed that the connections amongst cells and neurons in the retina and visual cortex for the purposes of perception and learning is not identical from one organism to another at birth; rather the connections appear random but with some spatially organised distribution of components. This may be uniform in the case of distributing different feature detectors across the retina, or it may show concentrations of certain types of cells such as the increased density of cones in the fovea and the corresponding small receptive fields of the ganglion cells. Furthermore, given the nature of the input signal a parallel system is used to optimally process the huge amount of signals received at the millions of photoreceptors. It is simply not the case that a single neuron processes all input data but that many neurons do so in parallel, connecting to varying numbers of input cells. Importantly, the number of neurons thought to be carrying out complex processing is far fewer in number than the size of the input. The neuron function is also assumed to largely be independent of most other neurons, although there may be some relations with those that are near.

The neural network connections between the input layer $l_0$ and neuron layer $l_1$ take the general idea of random connections into account so that relatively fewer neurons randomly sub-sample the entire input space with replacement. The motivation for random variable subsampling is to reduce overfitting by increasing the variance of what each neuron learns about the input data. The sampling is also carried out over a lognormal distribution so that smaller subsamples are biased, making it unlikely that anomalies will contaminate most samples. (It may be noted that the random subsampling has parallels with the use of `drop out' in Deep Learning where connections between layers of neurons are randomly dropped as a method of regularisation.) The neurons process the input data in parallel and independently of each other; each learning from its experience of the world. However, the prediction scores and binary decision are summed up at the output nodes to summarise the results of each computational unit. This not only has benefits for anomaly detection over an unordered data set but also for distributed processing applications.   

\item \emph{Immediate Intelligence:} 
In early exploration of the visual system the retina was assumed to be a simple data capturing device that relayed information to the brain. However, it was demonstrated by the works of \citet{Hartline1940THERF}, \citet{Kuffler1953DischargePA}, \citet{Barlow1953SummationAI} and especially that of \citet{LettvinFrog1959}, that the retina was already doing something intelligent. Barlow discovered that the eye provides the brain with information that is already, to a degree, organised and interpreted, instead of simply transmitting an image. While Lettvin's works suggested that a large part of the sensory machinery involved in a frog's feeding response may actually reside in the retina, with ganglion cells presumed to be simple `feature detectors' that responded immediately, quickly and reliably to certain stimuli related to the organism's environment. Thus, rather than the cells and neurons in the retina acting as conduits for the light stimuli, individual neurons perform intelligent operations and early visual processing is simply not just a mapping to higher regions in the brain that are less accessible. 

Inspired by some of the properties of ganglion cells the neurons in the $l_1$ layer perform computations that are not just a simple summation of values and thresholding with a non-linear activation function. Instead, intelligent decision making is already being carried out straightaway on the input values to decide which are considered meaningful (anomalies) parameter-free and without supervision. Indeed, each neuron first performs local anomaly detection using only its subsample to eject anomalies, and the learning is then carried out over the data that remain. The intention of such processing is to remove contaminants from the learning process. The neuron computations (initialisation, learning and prediction) are carried out on the immediate signal received and in real-time (no costly comparisons are carried out or the building of partition trees). Furthermore, viewing the individual neurons as simple but powerful computational units that act on the immediate signals received, and which work as quasi-independent units that each gather little pieces of evidence from small experiments to be combined by subsequent nodes, opens the door to consider more biologically inspired and plausible computational models of early sensory processing. 

\item \emph{Sparsity of activation and contrast invariance:}
In 1961, \citet{barlow1961possible} wrote a seminal article where he asked what the computational aims of the visual system are. He concluded that one of the main aims of visual processing is the reduction of redundancy. While the brightnesses of neighbouring points in images are usually very similar, the retina reduces this redundancy. Single neurons are regarded as the prime movers of these mechanisms with activities of neurons conjectured to quite simply be thought processes. His work was thus central to the field of statistics of natural scenes that relates the statistics of images of real world scenes to the properties of the nervous system. The study of neurons in the retina have shown that it is not the actual intensity values---corresponding to the number of photons arriving at the photoreceptors---that are important, but rather the specific and sparsely occurring contrasted patterns such as points, lines, bars and edges. Indeed, cells have been found to continue to respond to the same trigger feature in spite of changes in light intensity over many decades of investigations. Light is the agent, but it is the detailed pattern of light that carries the information---the overall illumination level is disregarded \cite{Maturana1960AnatomyAP}. 

Sparse coding is a natural consequence of building a neural network using the perception algorithm as individual components. Each neuron only fires sparsely in that there must be rarely occurring observations in its input subsample. Furthermore, the output nodes that sum up the signals also have a sparse representation in that the continuous valued sums and the binary value sums are rarely positive to indicate an anomaly. The network calculations also adhere to Wertheimer's principle of contrast invariance which states that the actual grey level values of an image are not important, rather it is the difference in intensity values that carries the essential information; and this is sparsely occurring in its nature. Thus, the network will continue to largely respond with the same output to the same essence of its stimulus regardless of the overall variation in the intensity of values calculated at the input layer. 

\item \emph{Activation under noise or constant stimuli:} 
Humans do not perceive structure or meaning in randomly distributed signals or where there is uniformity in value according to the Helmholtz principle. An example of the former is an ever changing uniformly random noise image of black and white pixels where it is not expected that a subject would perceive anything meaningful or be able to recall an image with clarity. An example of the latter is an image simply of one intensity or colour (although in this case it could be argued that a grouping of pixel intensities or colour is observed no further perception occurs). The neural network, and indeed the neuron models are designed such that they do not fire any excitatory outputs for such random or uniform input and this corresponds to the network not transmitting information to subsequent processes when patterns of data of relevance are not observed. The network perceives no meaningful structure that corresponds to unexpected observations and hence does not flag any as anomalies. 

\item \emph{Mass Action, Equipotentiality and Redundancy:} In neuroscience, the Mass Action principle suggests that the proportion of the brain that is injured is directly proportional to the decreased efficiency and ability of the processing. The related principle of Equipotentiality on the other hand suggests the apparent capacity of any functional part of the brain to carry out the functions which are lost by the destruction of other parts. Although areas of specific functionality have been found in the brain, this still may hold true within regions or for certain aspects of cognition. This suggests a universal algorithm or process amongst all or specific parts of the brain. In light of these principles it is interesting to note that the neural network is not reliant on any single node or pathway, rather its computations are distributed and the network is equally affected on average by any portion of neurons failing. This ensures that the parallel processing can continue to perform anomaly detection unto a gradual or critical breakdown of neurons and connections. Furthermore, in both the biological and distributed computing sense it also enables regeneration or replacement of failing nodes while the network continues to perform---with new neurons and nodes effectively taking over the roles. While this redundancy and resiliency may appear irrelevant to our model calculations on a single computing device, it is built into the network naturally so that for any distributed models or applications where there can be a loss of nodes, connections or the signal itself, the network can still continue to achieve its global task. Indeed, new nodes can be co-opted to perform the computations since each is essentially carrying out the same function but over different subsamples of the input data.

\end{enumerate}

%

\section{Experimental Results}
This section provides results of the neural network against the perception algorithm \cite{nassir2021anomaly}, the brilliantly performing isolation forest algorithm \cite{Liu2008} (sklearn implementation \cite{scikitlearn}), and the fast HBOS algorithm \cite{goldstein2012} (pyod implementation \cite{zhao2019pyod}). All algorithms are used with their default parameters and the experiments are carried out on a set of publicly available data sets provided by \citet{Rayana:2016} and \citet{credit_card_fraud2015}; the names and properties are shown in Table \ref{table:data_set_properties}. The chosen metric is the AUC-score due to its widespread use in evaluation and because it corresponds to how well anomalies are ranked at the top of a list (which is thought to be beneficial to end users). It is important to note however that this is not necessarily the best or most appropriate way to measure algorithm performance. Indeed, better methods need to be investigated in future. For additional comparison the F1-scores are also provided as a summary of the decision boundary resulting in precision and recall scores. The runtime is also given because it is an important factor in many applications and for usability. The results are shown in Tables \ref{table:auc_results}, \ref{table:f1_results}  and \ref{table:dataset_total_time} with the best performing algorithm for a given data set highlighted in bold.

The results clearly demonstrate the improvement---often significant---in the AUC-scores achieved by the neural network over the single neuron model (original perception algorithm) on almost all the data sets. The performance is however similar to the isolation forest algorithm and HBOS. In the case of the F1-scores however, the perception algorithm performs generally the best and often times by a considerable margin over the other algorithms. It gives a better decision boundary that balances the precision and recall scores better. The neural network---although it gives a higher recall than the perception algorithm---its precision is overly poorer which results in an overall lower F1-score. This aspect of the neural network requires improvement. Regarding the speed of the algorithms, the perception algorithm is usually fastest but similar to HBOS. However, the neural network takes considerably more time than both of these, but faster than isolation forest. This is simply due to running more neuron computations over the data which is currently done in serial fashion with only some basic python numpy optimisation. However, note that the neural network runs fast for many practical applications, even when the data set size runs into hundreds of thousands of examples due to its use of subsampling, and the underlying simplicity of the neuron computations.   

\begin{table}
\robustify\bfseries
\centering
\begin{tabular}{lrrr}
\toprule
       Name &  \# examples &  \# features &  \% anomalies \\
\midrule
       pima &          768 &            8 &         34.90 \\
credit-card &       284807 &           29 &          0.17 \\
     cardio &         1831 &           21 &          9.61 \\
    shuttle &        49097 &            9 &          7.15 \\
       musk &         3062 &          166 &          3.17 \\
       http &       567498 &            3 &          0.39 \\
       smtp &        95156 &            3 &          0.03 \\
    thyroid &         3772 &            6 &          2.47 \\
     lympho &          148 &           18 &          4.05 \\
        wbc &          378 &           30 &          5.56 \\
mammography &        11183 &            6 &          2.32 \\
      glass &          214 &            9 &          4.21 \\
 satimage-2 &         5803 &           36 &          1.22 \\
\bottomrule
\end{tabular}

  \caption{Names and properties of data sets selected for experiments that show the varying numbers of examples, features and percentages of anomalies.}
  \label{table:data_set_properties}
\end{table}

\begin{table}
\robustify\bfseries
\centering
\resizebox{\textwidth}{!}{%
\begin{tabular}{lllll}
\toprule
    Dataset &          HBOS & IsolationForest & NeuralNetwork &   Perception \\
\midrule
     cardio &          0.85 &   \textbf{0.93} &          0.92 &         0.77 \\
credit-card & \textbf{0.95} &   \textbf{0.95} & \textbf{0.95} &         0.93 \\
      glass &          0.71 &            0.71 & \textbf{0.74} &         0.58 \\
       http &          0.99 &    \textbf{1.0} &  \textbf{1.0} & \textbf{1.0} \\
     lympho &  \textbf{1.0} &    \textbf{1.0} &          0.99 &         0.95 \\
mammography &          0.83 &            0.85 & \textbf{0.89} &         0.72 \\
       musk &  \textbf{1.0} &    \textbf{1.0} &  \textbf{1.0} & \textbf{1.0} \\
       pima & \textbf{0.71} &            0.67 &          0.66 &         0.57 \\
 satimage-2 &          0.98 &   \textbf{0.99} &          0.97 &         0.93 \\
    shuttle &          0.98 &    \textbf{1.0} &          0.99 &         0.98 \\
       smtp &           0.8 &   \textbf{0.91} &           0.8 &          0.8 \\
    thyroid &          0.95 &   \textbf{0.98} &          0.94 &         0.86 \\
        wbc & \textbf{0.96} &            0.94 &          0.95 &         0.76 \\
\bottomrule
\end{tabular}

}
 \caption{AUC-scores. There is a general and significant improvement achieved by the neural network over the perception algorithm. However, the results of the network against Isolation Forest and HBOS are largely similar.}
  \label{table:auc_results}
\end{table}

\begin{table}
\robustify\bfseries
\centering
\resizebox{\textwidth}{!}{%
\begin{tabular}{lllll}
\toprule
    Dataset &           HBOS & IsolationForest &  NeuralNetwork &     Perception \\
\midrule
     cardio &          0.451 &           0.524 & \textbf{0.626} &          0.455 \\
credit-card &           0.03 &  \textbf{0.078} &          0.031 &          0.061 \\
      glass &          0.065 &  \textbf{0.125} &          0.065 &          0.095 \\
       http &          0.097 &           0.074 &           0.11 &  \textbf{0.26} \\
     lympho &          0.571 &           0.197 &          0.545 & \textbf{0.714} \\
mammography &          0.142 &           0.171 &          0.242 & \textbf{0.305} \\
       musk &           0.48 &            0.52 &          0.919 & \textbf{0.942} \\
       pima & \textbf{0.325} &           0.319 &          0.249 &          0.125 \\
 satimage-2 &          0.202 &           0.202 &          0.312 & \textbf{0.487} \\
    shuttle &          0.798 &           0.761 &          0.527 & \textbf{0.931} \\
       smtp &          0.005 &           0.004 &          0.008 & \textbf{0.012} \\
    thyroid &          0.323 &  \textbf{0.382} &          0.302 &          0.362 \\
        wbc &          0.542 &           0.536 &          0.405 &  \textbf{0.56} \\
\bottomrule
\end{tabular}

}
 \caption{F1-scores. The neural network has generally poorer F1-scores than the perception algorithm. This aspect of the network requires improvement where although it achieves higher recall in general than the perception algorithm, its precision is much lower, resulting in a lower F1-score.}
  \label{table:f1_results}
\end{table}

\begin{table}
\robustify\bfseries
\centering
\resizebox{\textwidth}{!}{%
\begin{tabular}{llrrl}
\toprule
    Dataset &           HBOS &  IsolationForest &  NeuralNetwork &     Perception \\
\midrule
     cardio &          0.008 &            0.718 &          0.299 & \textbf{0.007} \\
credit-card &          0.747 &           23.354 &         10.405 & \textbf{0.432} \\
      glass &          0.003 &            0.169 &          0.176 & \textbf{0.001} \\
       http & \textbf{0.203} &           27.128 &         12.659 &          0.635 \\
     lympho &          0.005 &            0.168 &          0.175 & \textbf{0.001} \\
mammography & \textbf{0.013} &            0.889 &          0.405 &          0.015 \\
       musk &          0.063 &            1.025 &          0.653 & \textbf{0.011} \\
       pima &          0.007 &            0.414 &          0.218 & \textbf{0.001} \\
 satimage-2 &          0.021 &            0.739 &          0.492 & \textbf{0.008} \\
    shuttle &  \textbf{0.05} &            4.603 &          1.127 &          0.063 \\
       smtp & \textbf{0.036} &            4.868 &          1.923 &          0.111 \\
    thyroid & \textbf{0.007} &            0.630 &          0.310 &          0.009 \\
        wbc &          0.011 &            0.535 &          0.261 & \textbf{0.001} \\
\bottomrule
\end{tabular}

}
  \caption{The total initialisation, training and prediction runtime in seconds by the anomaly detection methods. The perception algorithm and HBOS complete fast. The neural network (and isolation forest) is slower by comparison but would benefit from a parallelised implementation.} 
  \label{table:dataset_total_time}
\end{table}

%

\section{Conclusion}
\label{section:conclusion} 

The present work establishes practical connections between the approach taken by the anomaly detection algorithm of \citet{nassir2021anomaly}, and prior decades of research in neurophysiology and computational neuroscience. The algorithm is conceptualised as a neuron model that receives indicator streams at each of its receptor nodes which are summed over a unit time interval akin to counting photons by photoreceptors in the retina. The neuron learns a model of the world that it experiences and outputs scores and decisions for each of its inputs indicating whether they are anomalous or normal. The neuron performs a non-linear function for each of its inputs, is fast to compute, responds only to contrast, computes on immediate input in discrete time intervals, has adaptive thresholds, is parameter-free, is an internally reliable component, is stable in the presence of external noise and activates sparsely. Stacks of such computational units are constructed together to form a uni-directional neural network showing how anomaly detection over unordered data can be carried out as parallel neural computations. The neural network is practically parameter-free and processes in parallel units that can learn independently and fire in the presence of rarely occurring unusual values. It has properties of redundancy and global degradation, and robustness to data extremities that can mask lesser anomalies. The network also demonstrates that only random variable subsamples of the input data is required for accurate anomaly detection according to the AUC-score metric. The network contrasts with the plethora of autoencoders applied to anomaly detection where it operates in an unsupervised manner without feedback learning from the outputs, is parameter-free and without specification of thresholds. For larger data sets an important observation (as has been similarly pointed out by \citet{Liu2008}) is that all the data is not necessary to be used for optimal or satisfactory performance. Rather, for point anomaly detection only an approximately constant number of samples is required provided the random subsamples are taken over the entire data set.

The empirical results on a wide range of data sets are encouraging. This first neural network extended simply for multidimensional data, has a marked increase in AUC-score over the single neuron model (perception algorithm) and is competitive against the isolation forest and HBOS algorithms. However, its decision thresholds result in poorer F1-scores in general over the perception algorithm, where there is often high recall but low precision. Hence for making automated thresholded decisions the perception algorithm performs better, but the neural network provides better ranking. An end user using the results of these methods could choose one over the other depending on his requirements, or even take the best of both. Future work will aim at improving the neural network decision boundary. Furthermore, although the empirical results of the network on anomaly detection problems are promising a thorough evaluation on real-world production data is still required. I believe that real naturally occurring anomaly detection data sets will differentiate algorithm performance the best. With regards to speed the neural network is significantly slower than the perception algorithm due to the additional subsampling, computational units and connections. However, the parallel nature of the network construction signals the potential of implementing it in parallel computing architectures and thus improving the speed.

This neural network has taken inspiration from visual processes, and having conceptualised the perception algorithm as computational neurons and applied the network to the problem of anomaly detection over unordered data, the path to consider real-world signals such as those experienced by the human retina has been opened. Indeed, it will be intriguing to investigate the spatial organisation of the modifiable units and connections for the detection of low level features of visual scenes and other sense modalities---all with anomaly detection as the primary functional aim.

%

\section{Acknowledgements}
\label{section:acknowledgements} 
This research was funded by Endeavr Wales and Airbus.

\bibliography{Bibliography} 

\begin{thebibliography}{30}
\providecommand{\natexlab}[1]{#1}
\providecommand{\url}[1]{\texttt{#1}}
\expandafter\ifx\csname urlstyle\endcsname\relax
  \providecommand{\doi}[1]{doi: #1}\else
  \providecommand{\doi}{doi: \begingroup \urlstyle{rm}\Url}\fi

\bibitem[Aggarwal(2017)]{AggarwalOutlierAnalysis2017}
Charu~C. Aggarwal.
\newblock \emph{Outlier Analysis}.
\newblock Springer Publishing Company, Incorporated, 2nd edition, 2017.
\newblock ISBN 3319475770.

\bibitem[Barlow(1953)]{Barlow1953SummationAI}
H.~B. Barlow.
\newblock Summation and inhibition in the frog's retina.
\newblock \emph{The Journal of Physiology}, 119, 1953.

\bibitem[Barlow(1961)]{barlow1961possible}
H.~B. Barlow.
\newblock Possible principles underlying the transformation of sensory
  messages.
\newblock \emph{Sensory communication}, 1\penalty0 (01), 1961.

\bibitem[Barlow(1972)]{Barlow72}
H.~B. Barlow.
\newblock Single units and sensation: A neuron doctrine for perceptual
  psychology?
\newblock \emph{Perception}, 1\penalty0 (4):\penalty0 371--394, 1972.

\bibitem[Barnett and Lewis(1978)]{Barnett1978}
V.~Barnett and T.~Lewis.
\newblock \emph{Outliers in statistical data.}
\newblock John Wiley \& Sons Ltd., 2nd edition edition, 1978.

\bibitem[Chen et~al.(2017)Chen, Sathe, Aggarwal, and Turaga]{Chen2017OutlierDW}
J.~Chen, Saket~K. Sathe, Charu~C. Aggarwal, and Deepak~S. Turaga.
\newblock Outlier detection with autoencoder ensembles.
\newblock In \emph{SDM}, 2017.

\bibitem[Desolneux et~al.(2007)Desolneux, Moisan, and Morel]{Morel07}
Agnès Desolneux, Lionel Moisan, and Jean-Michel Morel.
\newblock \emph{From Gestalt Theory to Image Analysis: A Probabilistic
  Approach}.
\newblock Springer Publishing Company, Incorporated, 1st edition, 2007.
\newblock ISBN 0387726357.

\bibitem[Goldstein and Dengel(2012)]{goldstein2012}
Markus Goldstein and Andreas Dengel.
\newblock Histogram-based outlier score (hbos): A fast unsupervised anomaly
  detection algorithm.
\newblock 2012.

\bibitem[Goldstein and Uchida(2016)]{Goldstein2016}
Markus Goldstein and Seiichi Uchida.
\newblock A comparative evaluation of unsupervised anomaly detection algorithms
  for multivariate data.
\newblock \emph{PLoS One}, 11\penalty0 (4), 4 2016.
\newblock ISSN 1932-6203.
\newblock \doi{10.1371/journal.pone.0152173}.

\bibitem[Hampel(2001)]{Hampel2001}
Frank Hampel.
\newblock Robust statistics: a brief introduction and overview.
\newblock \emph{Seminar for Statistics, ETH Zurich, Switzerland}, 2001.

\bibitem[Hartline(1940)]{Hartline1940THERF}
Haldan~Keffer Hartline.
\newblock The receptive fields of optic nerve fibers.
\newblock \emph{American Journal of Physiology}, 130:\penalty0 690--699, 1940.

\bibitem[Hawkins et~al.(2002)Hawkins, He, Williams, and
  Baxter]{Hawkins02outlierdetection}
Simon Hawkins, Hongxing He, Graham Williams, and Rohan Baxter.
\newblock Outlier detection using replicator neural networks.
\newblock In \emph{Proc. of the Fifth Int. Conf. on Data Warehousing and
  Knowledge Discovery (DaWaK02}, pages 170--180, 2002.

\bibitem[Hubel and Wiesel(1959)]{Hubel1959ReceptiveFO}
D.~Hubel and T.~Wiesel.
\newblock Receptive fields of single neurones in the cat's striate cortex.
\newblock \emph{The Journal of Physiology}, 148, 1959.

\bibitem[Kingma and Welling(2014)]{Kingma2014AutoEncodingVB}
Diederik~P. Kingma and Max Welling.
\newblock Auto-encoding variational bayes.
\newblock \emph{CoRR}, abs/1312.6114, 2014.

\bibitem[Kuffler(1953)]{Kuffler1953DischargePA}
Stephen~W. Kuffler.
\newblock Discharge patterns and functional organization of mammalian retina.
\newblock \emph{Journal of neurophysiology}, 16 1:\penalty0 37--68, 1953.

\bibitem[Lettvin et~al.(1959)Lettvin, Maturana, McCulloch, and
  Pitts]{LettvinFrog1959}
J.~Y. Lettvin, H.~R. Maturana, W.~S. McCulloch, and W.~H. Pitts.
\newblock What the frog's eye tells the frog's brain.
\newblock \emph{Proceedings of the IRE}, 47\penalty0 (11):\penalty0 1940--1951,
  1959.

\bibitem[Liu et~al.(2008)Liu, Ting, and Zhou]{Liu2008}
Fei~Tony Liu, Kai~Ming Ting, and Zhi-Hua Zhou.
\newblock Isolation forest.
\newblock In \emph{Proceedings of the 2008 Eighth IEEE International Conference
  on Data Mining}, ICDM, pages 413--422, Washington, DC, USA, 2008. IEEE
  Computer Society.

\bibitem[Marr(1982)]{Mar82}
David Marr.
\newblock \emph{Vision: A Computational Investigation into the Human
  Representation and Processing of Visual Information}.
\newblock Henry Holt and Co., Inc., USA, 1982.
\newblock ISBN 0716715678.

\bibitem[Maturana et~al.(1960)Maturana, Lettvin, McCulloch, and
  Pitts]{Maturana1960AnatomyAP}
Humberto~R. Maturana, Jerome~Y. Lettvin, Warren~S. McCulloch, and Walter Pitts.
\newblock Anatomy and physiology of vision in the frog (rana pipiens).
\newblock \emph{The Journal of General Physiology}, 43:\penalty0 129 -- 175,
  1960.

\bibitem[McCulloch and Pitts(1943)]{mcculloch43a}
Warren McCulloch and Walter Pitts.
\newblock A logical calculus of ideas immanent in nervous activity.
\newblock \emph{Bulletin of Mathematical Biophysics}, 5:\penalty0 127--147,
  1943.

\bibitem[Mohammad(2021)]{nassir2021anomaly}
Nassir Mohammad.
\newblock Anomaly detection using principles of human perception, 2021.

\bibitem[Ng(2012)]{NgML}
Andrew Ng.
\newblock \emph{Machine learning: Programming Exercise 8: Anomaly Detection and
  Recommender Systems}.
\newblock 2012.

\bibitem[Pedregosa et~al.(2011)Pedregosa, Varoquaux, Gramfort, Michel, Thirion,
  Grisel, Blondel, Prettenhofer, Weiss, Dubourg, Vanderplas, Passos,
  Cournapeau, Brucher, Perrot, and Duchesnay]{scikitlearn}
F.~Pedregosa, G.~Varoquaux, A.~Gramfort, V.~Michel, B.~Thirion, O.~Grisel,
  M.~Blondel, P.~Prettenhofer, R.~Weiss, V.~Dubourg, J.~Vanderplas, A.~Passos,
  D.~Cournapeau, M.~Brucher, M.~Perrot, and E.~Duchesnay.
\newblock Scikit-learn: Machine learning in {P}ython.
\newblock \emph{Journal of Machine Learning Research}, 12:\penalty0 2825--2830,
  2011.

\bibitem[Pozzolo et~al.(2015)Pozzolo, Caelen, Johnson, and
  Bontempi]{credit_card_fraud2015}
Andrea~Dal Pozzolo, Olivier Caelen, Reid~A. Johnson, and Gianluca Bontempi.
\newblock Calibrating probability with undersampling for unbalanced
  classification.
\newblock In \emph{2015 IEEE Symposium Series on Computational Intelligence},
  pages 159--166, 2015.

\bibitem[Ranzato et~al.(2006)Ranzato, Poultney, Chopra, and Lecun]{Ranzato2006}
Marc'Aurelio Ranzato, Christopher Poultney, Sumit Chopra, and Yann Lecun.
\newblock Efficient learning of sparse representations with an energy-based
  model.
\newblock 2006.

\bibitem[Rayana(2016)]{Rayana:2016}
Shebuti Rayana.
\newblock Odds library.
\newblock Stony Brook University, Department of Computer Sciences, 2016.
\newblock URL \url{http://odds.cs.stonybrook.edu}.

\bibitem[Rifai et~al.(2011)Rifai, Vincent, Muller, Glorot, and
  Bengio]{Rifai_contractive_2011}
Salah Rifai, Pascal Vincent, Xavier Muller, Xavier Glorot, and Yoshua Bengio.
\newblock Contractive auto-encoders: Explicit invariance during feature
  extraction.
\newblock In \emph{Proceedings of the 28th International Conference on
  International Conference on Machine Learning}, ICML'11, page 833–840,
  Madison, WI, USA, 2011. Omnipress.
\newblock ISBN 9781450306195.

\bibitem[Rosenblatt(1958)]{rosenblatt1958perceptron}
F.~Rosenblatt.
\newblock The perceptron: A probabilistic model for information storage and
  organization in the brain.
\newblock \emph{Psychological Review}, 65\penalty0 (6):\penalty0 386--408,
  1958.
\newblock ISSN 0033-295X.
\newblock \doi{10.1037/h0042519}.

\bibitem[Vincent et~al.(2008)Vincent, Larochelle, Bengio, and
  Manzagol]{Vincent_denoising_2008}
Pascal Vincent, Hugo Larochelle, Yoshua Bengio, and Pierre-Antoine Manzagol.
\newblock Extracting and composing robust features with denoising autoencoders.
\newblock In \emph{Proceedings of the 25th International Conference on Machine
  Learning}, ICML '08, page 1096–1103, New York, NY, USA, 2008. Association
  for Computing Machinery.
\newblock ISBN 9781605582054.
\newblock \doi{10.1145/1390156.1390294}.
\newblock URL \url{https://doi.org/10.1145/1390156.1390294}.

\bibitem[Zhao et~al.(2019)Zhao, Nasrullah, and Li]{zhao2019pyod}
Yue Zhao, Zain Nasrullah, and Zheng Li.
\newblock Pyod: A python toolbox for scalable outlier detection.
\newblock \emph{Journal of Machine Learning Research}, 20\penalty0
  (96):\penalty0 1--7, 2019.
\newblock URL \url{http://jmlr.org/papers/v20/19-011.html}.

\end{thebibliography}
\bibliographystyle{plainnat}

\end{document}